\documentclass[journal]{IEEEtran}

\usepackage{times}
\usepackage{epsfig}
\usepackage{graphicx}
\usepackage{amsmath}
\usepackage{amssymb}
\graphicspath{ {./images/}}

\ifCLASSOPTIONcompsoc
    \usepackage[caption=false, font=normalsize, labelfont=sf, textfont=sf]{subfig}
\else
    \usepackage[caption=false, font=footnotesize]{subfig}
\fi
\usepackage{multirow}
\usepackage{adjustbox}
\usepackage{booktabs}
\usepackage[table,xcdraw]{xcolor}
\newcommand{\hyperfootnote}[1]{
    \footnote{\href{#1}{#1}}
}

\usepackage[pagebackref=true,breaklinks=true,colorlinks,bookmarks=false]{hyperref}
\usepackage[capitalise]{cleveref}
\usepackage{cite}
\begin{document}
%
\title{Moving Towards Centers: Re-ranking with Attention and Memory for Re-identification}
%
%
%

\author{Yunhao~Zhou,
        Yi~Wang,~\IEEEmembership{Member,~IEEE,}
        and~Lap-Pui~Chau,~\IEEEmembership{Fellow,~IEEE}
\thanks{Yunhao Zhou, Yi Wang, and Lap-Pui Chau are with School of Electrical
and Electronics Engineering, Nanyang Technological University, Singapore,
639798 (e-mail: zh0022ao@e.ntu.edu.sg, wang1241@e.ntu.edu.sg, elpchau@ntu.edu.sg).
Corresponding author: Lap-Pui Chau.}
}

\markboth{Journal of \LaTeX\ }%
{Shell \MakeLowercase{\textit{et al.}}: Bare Demo of IEEEtran.cls for IEEE Journals}
%



\maketitle

\begin{abstract}
Re-ranking utilizes contextual information to optimize the initial ranking list of person or vehicle re-identification (re-ID), which boosts the retrieval performance at post-processing steps.
This paper proposes a re-ranking network to predict the correlations between the probe and top-ranked neighbor samples.
Specifically, all the feature embeddings of query and gallery images are expanded and enhanced by a linear combination of their neighbors, with the correlation prediction serving as discriminative combination weights.
The combination process is equivalent to moving independent embeddings toward the identity centers, improving cluster compactness.
For correlation prediction, we first aggregate the contextual information for probe's $k$-nearest neighbors via the Transformer encoder.
Then, we distill and refine the probe-related features into the Contextual Memory cell via attention mechanism.
Like humans that retrieve images by not only considering probe images but also memorizing the retrieved ones, the Contextual Memory produces multi-view descriptions for each instance.
Finally, the neighbors are reconstructed with features fetched from the Contextual Memory, and a binary classifier predicts their correlations with the probe.
Experiments on six widely-used person and vehicle re-ID benchmarks demonstrate the effectiveness of the proposed method. 
Especially, our method surpasses the state-of-the-art re-ranking approaches on large-scale datasets by a significant margin, i.e., with an average 4.83\% CMC@1 and 14.83\% mAP improvements on VERI-Wild, MSMT17, and VehicleID datasets.
\end{abstract}


%
\IEEEpeerreviewmaketitle

\section{Introduction}
\begin{figure}[tb]
    \centering
    \includegraphics[page=8, width=\columnwidth]{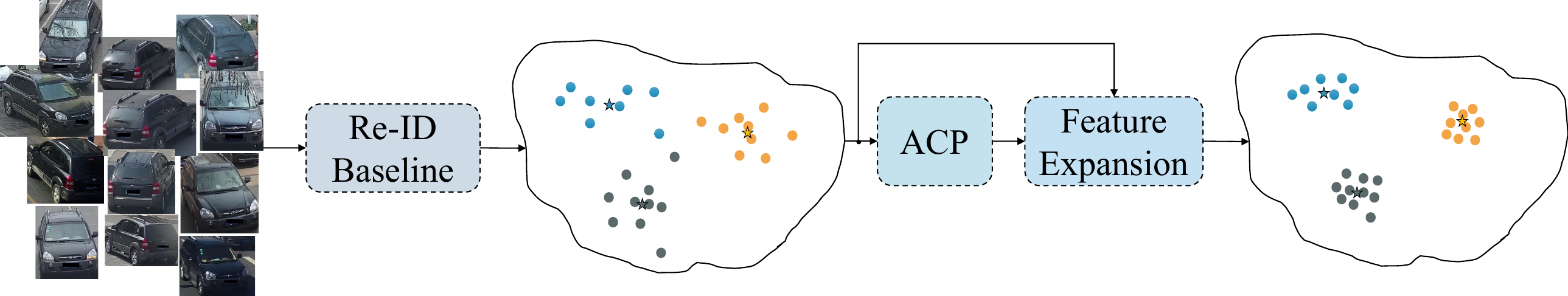}
    \caption{General pipeline of re-ID and our Attention-based Correlation Predictor (ACP). Embeddings and identities are shown with dots and colors.
    Our re-ranking method aggregates each embedding and the corresponding $k$-nearest neighbors with predicted correlations, which moves independent embeddings towards the identity centers marked with stars.}
    \label{fig:Re-ID baseline and moving towards centers}
\end{figure}

Recently, re-identification (re-ID) tasks have drawn increasing interest in the computer vision society. 
Given an image from the query set, re-ID aims at finding all images containing the same instance as the query across a large gallery image set. 
In general, query and gallery images are captured by different cameras in multiple scenes. 
Thus, re-ID is often considered a sub-problem of image retrieval at the instance-specific level. 
Re-ID has a wide variety of applications according to the target instances. 
For example, person re-ID helps the criminal investigation \cite{ye2016person} by looking for suspected persons with city surveillance cameras. 
Vehicle re-ID can analyze traffic \cite{zheng2020vehiclenet}, which is a crucial part of intelligent transportation systems in smart cities. 
The problem setting that retrieving instances from multiple non-overlapping cameras brings many challenges, such as varying viewpoints, illumination changes, occlusions, low-image resolutions, and cluttered backgrounds. 
Therefore, the key to re-ID is building robust and discriminative feature embeddings to minimize intra-identity distance and maximize inter-identity discrepancy. 
As shown in \cref{fig:Re-ID baseline and moving towards centers}, a re-ID baseline maps images to an embedding space.
Early research mainly focused on designing hand-crafted image descriptors \cite{xiong2014person, matsukawa2016hierarchical}. 
As more and more training data becomes available, the ability of Convolution Neural Networks (CNN) to learn robust feature representations from data pushes the re-ID performance to a new level \cite{luo2019strong, khorramshahi2020devil, zhou2018vehicle}.

Basic re-ID models retrieve instances with pairwise distance measure, which only considers individual characters between two separate images at a time.
Performance degrades quickly in challenging scenarios where images containing the same instance cannot be embedded into a small cluster and are not close enough to each other. 
To overcome the drawbacks of the pairwise matching rule, re-ranking leverages the contextual information in local neighbors to optimize the initial ranking list of re-ID, which conducts retrieval by integrating information from multiple images.
For example, \cite{chum2007total, radenovic2018fine, arandjelovic2012three} aggregates the contextual information via directly averaging local neighboring embeddings.
It is equivalent to substituting the pairwise distance with the distance between centers of different groups of embeddings.
Some methods resort to more complex rules like common nearest neighbors \cite{ye2015coupled} or $k$-reciprocal neighbors \cite{zhong2017re}.
Re-ranking has been integrated into many re-ID methods \cite{sun2018beyond, luo2019strong, he2020fastreid} because of the performance boost it brings, especially on mean Average Precision (mAP).
However, most existing re-ranking approaches rely on hand-designed algorithms to leverage contextual information.
They lack the ability to handle complex neighborhood relationships, which yields unsatisfying results on large-scale datasets. 
Moreover, some re-ranking methods are time-consuming. 
For example, the $k$-reciprocal re-ranking \cite{zhong2017re} calculates $k$-reciprocal neighbors of each probe image to form a new Jaccard distance. 
Experiments on the largest test set of VERI-Wild show that it takes more than 1000s and 150GB RAM to finish re-ranking.

Given an initial ranking list from a re-ID baseline, humans achieve re-ranking by gradually verifying the retrieved images and memorizing the appearance information of the verified confident samples.
Our brain can summarize the scattered contextual information to form an overall description of a person or vehicle.
Based on this observation, we propose a Contextual Memory cell to fulfill the information aggregation purpose powered by attention mechanism.
We later reconstruct top-ranked samples by comparing their features with the refined Contextual Memory.
Because falsely retrieved samples share less common features with the memory, their reconstruction results will be different from those true matches, which can be used for correlation prediction, i.e., whether a retrieved sample is a true match for the probe image.
With the predicted correlations, we can shrink the independent feature embeddings to the identity centers through a weighted combination of their local neighbors in the original embedding space.
An overview of our re-ranking model is visualized in \cref{fig:Re-ID baseline and moving towards centers} where the improved cluster compactness can reduce false matches.
Specifically, we reformulate the re-ranking as a binary classification problem of the top-$k$ retrieved samples and propose an Attention-based Correlation Predictor (ACP) to predict the correlations. 
The correlation prediction consists of three key steps.
In the first step, each probe's $k$-nearest neighbors are fed into a Transformer encoder to aggregate the contextual information. 
Second, a Contextual Memory initialized by attention mechanism distills the probe-related contextual features.
We further refine the memory with a small group of highly confident samples for the purpose of eliminating feature pollutions brought by interfering false matches.
Third, we reconstruct neighbor embeddings with features fetched from the memory cell through Multi-Head Attention.
A binary classifier with sigmoid activation predicts the correlation between each reconstructed neighbor embedding and the probe.

The proposed method is evaluated on six widely used benchmarks including VERI-Wild \cite{VeRiWild}, MSMT17 \cite{MSMT17}, VehicleID \cite{VehicleID}, VeRi \cite{VeRi}, Market1501 \cite{Market1501} and DukeMTMC \cite{DukeMTMC}. 
Experimental results show that our method surpasses state-of-the-art re-ranking methods by a large margin on large-scale benchmarks VERI-Wild, MSMT17, and VehicleID.
For smaller benchmarks with limited training images, we ranked high amongst the competing methods.
Ablation study verifies the effectiveness of our Contextual Memory and refinement process. 
In summary, the contribution of this paper is threefold:
\begin{itemize}
\item We propose to use the Contextual Memory cell to mimic the remembering process that humans adopt for re-ID and re-ranking. Embeddings obtained from single-view images are expanded with attention mechanism, which learns to focus on the most discriminative regions.
\item We reformulate re-ranking as a binary classification problem. The independent embeddings are moved to the identity centers by combining local neighbors with correlation prediction produced by the classifier.
\item Extensive experiments on six re-ID datasets demonstrate the superiority of the proposed method. Detailed model studies reflect how each module and parameter affect the performance.
\end{itemize}

\section{Related Work}
\subsection{Deep feature representation of re-ID}
Recently, CNN-based feature representation learning has achieved great success on re-ID tasks.
In the early years, many re-ID methods \cite{zheng2017person, qian2017multi} with deep learning techniques directly learn a global feature embedding from the whole image.
However, a single global feature often fails to distinguish two similar-looking instances like vehicles in the same model or persons dressed in the same color and style.
To solve this problem, some methods extract supplementary features like strong discriminative regions \cite{khorramshahi2020devil}, viewpoints \cite{zhou2018vehicle, sarfraz2018pose} or attribute characteristics \cite{wang2020attribute} to provide auxiliary cues for the global feature.
For example, the Part-based Convolutional Baseline (PCB) \cite{sun2018beyond} uniformly partitions the learned feature map into multiple horizontal stripes.
Strip features provide part-level information from different body regions.
Zhou \textit{et al.} \cite{zhou2018vehicle} combined CNN with Long Short-Term Memory (LSTM) to learn the transformations across different viewpoints of vehicles. Multi-view vehicle representation can be inferred from a single view image input. 
Instead of focusing on model architectures, another line of work adopts metric learning methods to improve the feature discriminability. 
Hermans \textit{et al.} \cite{hermans2017defense} proposed a variant of triplet loss with batch hard sampling strategy to pull positive samples together and push negative samples away. 
It has become one of the most common choices \cite{luo2019strong, khorramshahi2020devil} to train deep re-ID networks supervised by the combination of triplet loss and cross-entropy identity loss. 

\subsection{Re-ranking for re-ID and Image Retrieval}
Re-ranking plays a crucial role in improving retrieval performance at post-processing steps. 
Given an initial ranking list, re-ranking refines the result utilizing contextual information in top-ranked samples. 
Although feature representation learning ushers in the blossom of deep neural networks, most existing re-ranking methods still stagnate in hand-designed rules when analyzing the relations between neighboring embeddings.
Chum \textit{et al.} \cite{chum2007total} proposed Average Query Expansion (AQE) for image retrieval tasks, which replaces query feature embeddings with the mean of top-$k$ retrieved gallery samples. 
Instead of taking the mean average, Radenovi{\'c} \textit{et al.} \cite{radenovic2018fine} resort weighted average named alpha Query Expansion ($\alpha$QE). 
The weights are calculated by taking the power of similarity between the query and top-$k$ retrieved samples with an exponent as a hyper-parameter $\alpha$. 
Because of its simplicity and robustness, $\alpha$QE has been adopted as a common approach of boosting retrieval performance by a number of image retrieval and re-ID works. 
To improve the discriminability, Arandjelovi{\'c} \textit{et al.} \cite{arandjelovic2012three} proposed Discriminate Query Expansion (DQE) which trains a linear Support Vector Machine (SVM) with top-ranked and bottom-ranked samples as positive and negative samples, respectively. 
The distances between samples and the decision boundary work as pseudo labels to aggregate the $k$-nearest neighbors. 
Direct utilization of top-ranked samples faces the challenge of false-match pollution. 
Qin \textit{et al.} \cite{qin2011hello} first proposed $k$-reciprocal nearest neighbors to eliminate outliers. 
Two images are called $k$-reciprocal nearest neighbors if they both ranked top-$k$ when the other image serves as a probe.
Bai and Bai \cite{bai2016sparse} proposed Sparse Contextual Activation (SCA) which encodes the local distribution of an image in contextual space.
Zhong \textit{et al.} \cite{zhong2017re} calculate the expanded $k$-reciprocal nearest neighbor sets whose Jaccard distance is aggregated with the original distance via convex combination.
Expanded Cross Neighbor (ECN) is introduced by \cite{sarfraz2018pose} which sums the distances of images in expanded neighbors.
Yu \textit{et al.} \cite{yu2017divide} divide the extracted features into multiple sub-features, then the contextual information is iteratively encoded and fused into new feature vectors. 
Ye \textit{et al.} \cite{ye2016person} proposed to consider not only the similarity of top-$k$ samples but also the dissimilarity of bottom-$k$ samples from different baseline methods.
Recently, Wang \textit{et al.} \cite{wang2019incremental} proposed reciprocal optimization that takes multiple queries into account.

The continuous progress of re-ranking methods pushes the performance forward by designing more and more sophisticated algorithms when exploiting the contextual information hiding in $k$-nearest neighbors. 
Hand-designed rules generalize well on small benchmarks but show difficulties in fitting large amounts of data.
Liu \textit{et al.} \cite{liu2021prgcn} proposed to use graph convolutional network for link probability prediction and replace the original Euclidean distance with the predicted probability.
Instead, we consider our correlation prediction as combination weights to shrink independent embeddings toward their identity centers. 
Therefore, we are not suffered from predicting the entire pairwise correlations between queries and galleries.
Besides, our method only requires finding the first-order neighbors for each embedding. 
The whole architecture can be easily implemented under existing deep learning frameworks with GPU acceleration available painlessly.

\begin{figure}[tb]
    \centering
    \includegraphics[page=7, width=\columnwidth]{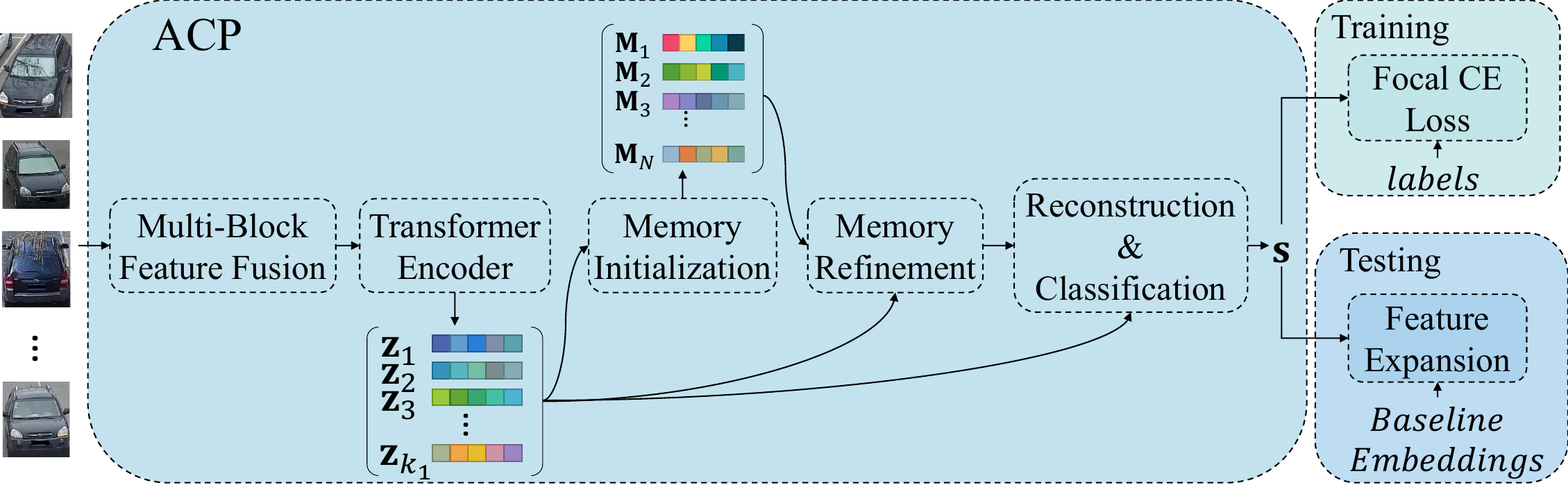}
    \caption{Architecture of our Attention-based Correlation Predictor.}
    \label{fig:ACP}
\end{figure}

\subsection{Attention and Transformer}
Transformers were introduced by \cite{vaswani2017attention} and have achieved huge success on a wide range of natural language processing tasks. 
The core idea of Transformer is to update the sequence with information aggregated from the entire input sequence via attention mechanism which captures complex relationships between different input tokens.
Recently, Transformer started to shine in computer vision society. 
Wang \textit{et al.} \cite{wang2018non} proposed Non-local Neural Networks that use self-attention to aggregate image feature representations from non-local regions.
Parmar \textit{et al.} \cite{parmar2018image} leverage the self-attention of Transformer for auto-regressive image generation. 
Carion \textit{et al.} \cite{carion2020end} proposed Detection Transformer (DETR) with self-attention layers built on top of convolutional backbones. 
Dosovitskiy \textit{et al.} \cite{dosovitskiy2020image} proved Vision Transformer (ViT) can achieve state-of-the-art performance on image recognition tasks, which relies on self-attention with flattened image patches as inputs discarding the convolutional architecture entirely. 
For re-ID, TransReID proposed by \cite{he2021transreid} shows that the self-attention between image patches can produce more robust features than CNN because information loss on details is avoided by removing convolution and downsampling operators.
In this work, the self-attention of Transformer is used to explore the contextual information in local neighborhoods.

\section{Our Approch}
In this section, we first formulate the re-ID baseline and re-ranking with feature expansion in \cref{sec:Overview of re-ID and re-ranking}. 
Next, the detailed model architecture is described in \cref{sec:Model Architecture for Re-ranking}.
We visualize the pipeline of our re-ranking approach, named Attention-based Correlation Predictor (ACP), in \cref{fig:ACP}.

\subsection{Overview of re-ID and re-ranking} \label{sec:Overview of re-ID and re-ranking}
Given an image from the query set, re-ID aims at retrieving images containing the same instance in the gallery set. 
Re-ID first maps images to $D$-dimensional vectors through a feature extractor $\phi$. 
We define the feature representation of images as $\mathbf{f} = \phi(\mathcal{I})$. 
The output feature vectors form two matrices: $\mathcal{Q} \in \mathcal{R}^{M \times D}$ and $\mathcal{G} \in \mathcal{R}^{N \times D}$ where $M$ and $N$ refer to the number of images in query and gallery sets, respectively. 
Then, pairwise distances between the feature representations in $\mathcal{Q}$ and $\mathcal{G}$ are calculated. 
The retrieval process of re-ID is achieved by sorting the distances in ascending order. 
It is expected that images holding the same instances will group together in the embedding space, so the correct matches in the gallery set will be ranked closer to the query.

\begin{figure}[tb]
    \centering
    \subfloat[Baseline embeddings]{
    \begin{minipage}[tb]{0.48\columnwidth}
        \begin{minipage}[tb]{\columnwidth}
            \centering
            \includegraphics[page=1, width=\columnwidth]{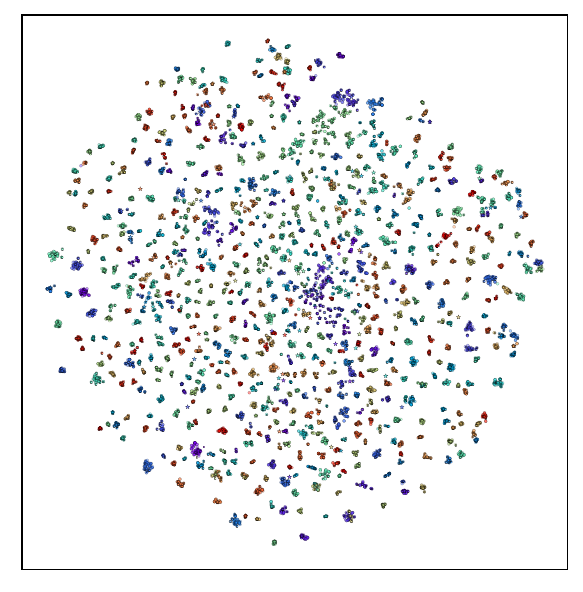}
        \end{minipage}
    \end{minipage}}
    \subfloat[Expanded embeddings]{
    \begin{minipage}[tb]{0.48\columnwidth}
        \begin{minipage}[tb]{\columnwidth}
            \centering
            \includegraphics[page=1, width=\columnwidth]{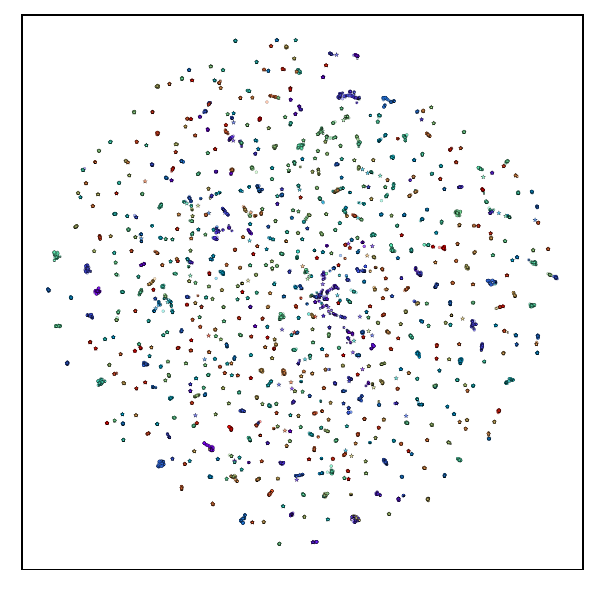}
        \end{minipage}
    \end{minipage}}
    \caption{Feature embedding visualization with t-SNE dimension reduction for Market1501.}
    \label{fig:visEffect}
\end{figure}

Our re-ranking network exploits the contextual information in top-ranked samples to optimize the initial ranking list produced by the re-ID baseline.
Given an image, we expand its embedding $\mathbf{p}$ with the linear combination of its $k_1$-nearest neighbors (including itself).
The weights are predicted by our ACP as shown in \cref{fig:ACP}.
The expansion moves independent embeddings toward their identity centers, which boosts the performance because of the improved cluster compactness.
Denote the embeddings of $k_1$-nearest neighbors of $\mathbf{p}$ as $\mathbf{N}_\mathbf{f} = \{\mathbf{f}_i\}_{i=1}^{k_1}$, and the corresponding original images as $\mathbf{N}_\mathbf{I} = \{\mathbf{I}_i\}_{i=1}^{k_1}$.
The feature expansion of $\mathbf{p}$ is, 
\begin{equation}
    \mathbf{p}^{\star} = \sum_{j=1}^{k_1}{\mathbf{N}_\mathbf{f}^{j} \odot \mathbf{s}_j}
\end{equation}
where $\mathbf{s} = {\rm ACP}(\mathbf{N}_\mathbf{I})$, $\mathbf{s} \in \mathcal{R}^{k_1}$ and $\odot$ is element-wise product.
Note that we perform feature expansion for both the query and gallery sets.
Finally, the pairwise distances between updated embeddings in two sets are re-calculated and sorted to accomplish the instance re-ID goal.
To give an intuitive perception of how the embedding moves, we visualize the distribution of extracted feature vectors before and after feature expansion in \cref{fig:visEffect}.
Specifically, the high-dimensional feature vectors (2048 dimensions) are mapped to 2-dimensional vectors by t-SNE \cite{van2008visualizing}.
It is obvious that features after expansion form tighter groups than the original ones.

\subsection{Model Architecture for Re-ranking} \label{sec:Model Architecture for Re-ranking}
Given an image, we sort its $k_1$-nearest neighbors into a sequence.
The attention is a powerful tool in aggregating contextual information over the sequence, while Contextual Memory can distill the probe-related features for correlation prediction, which are integrated into the ACP.
In this section, we first describe a multi-block feature fusion module. 
Then, we elaborate on how Contextual Memory is initialized and how the correlations are predicted via attention and the memory modules.

\subsubsection{Multi-Block Feature Fusion}
\begin{figure}[tb]
    \centering
    \includegraphics[page=2, width=0.96\columnwidth]{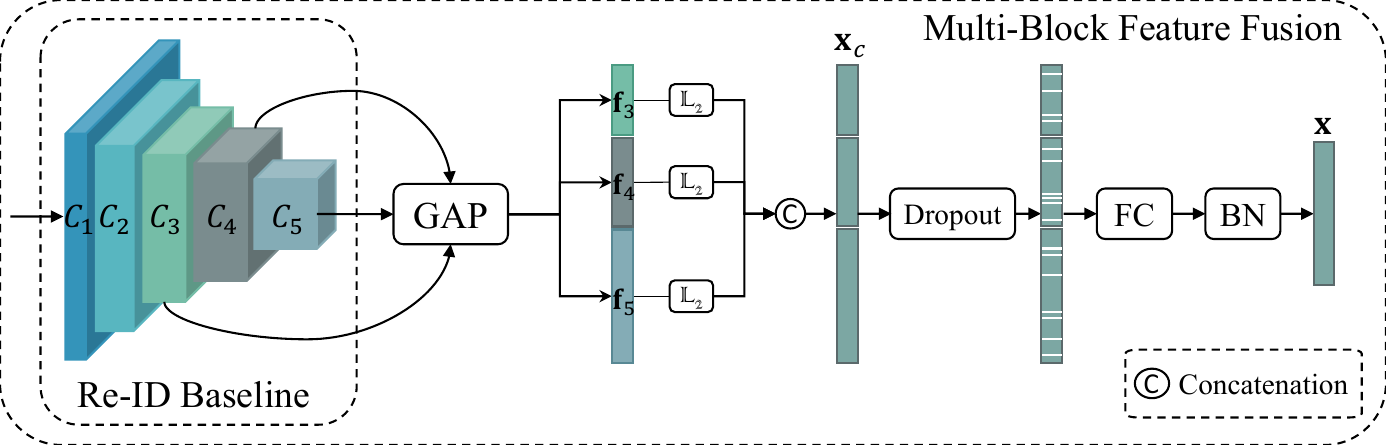}
    \caption{
    Multi-Block Feature Fusion. 
    The colored cuboids represent features from different blocks of the backbone ResNet-50 \cite{he2016deep}. 
    GAP is a global average pooling layer. 
    FC and BN are fully connected layer and Batch Normalization layer, respectively.}
    \label{fig:feature fusion}
\end{figure}

We visualize the feature fusion module in \cref{fig:feature fusion}.
The feature extractor of the re-ID baseline is ResNet-50 \cite{he2016deep} which consists of 5 blocks, i.e., $C_1$, $C_2$, $C_3$, $C_4$, and $C_5$. 
Feature maps from successive two blocks have a stride difference of 2.
The re-ID baseline model obtains image embeddings by feeding $C_5$ through a global average pooling (GAP) layer to reduce the spatial dimension. The features from shallower layers are discarded.
Some previous works leverage that lower-level information from shallower blocks to improve the re-ID performance. 
For example, \cite{chang2018multi} proposed a Factor Module (FM) and a Factor Selection Module (FSM) to extract and select features of details, respectively. 
\cite{wang2018resource} combines pooled feature vectors from different layers with a learnable combination weight.
To reduce the computation costs brought by stacking multiple layers of FMs, we design a multi-block feature fusion module similar to \cite{wang2018resource}.

Suppose the multi-block features for an image $\mathcal{I}$ as $\phi(\mathcal{I})= \{\mathbf{f}_1, \mathbf{f}_2, ..., \mathbf{f}_5 \}$, where $\mathbf{f}_j$ is $j$-th feature vector generated by performing global average pooling on CNN feature maps. 
For ResNet-50, we fuse the last three pooled feature vectors, i.e., $\{\mathbf{f}_3, \mathbf{f}_{4}, \mathbf{f}_5 \}$. 
The feature vectors are first normalized by $L_2$ normalization and rescaled with a learnable scale parameter. 
Instead of fusing featues with learned combination weights, our learnable parameter and re-scaling operation are inspired by \cite{ioffe2015batch}, which restores the expressive power of neural networks.
The scaled $L_2$ normalization layer is formulated as,
\begin{equation}
    \hat{\mathbf{f}_{j}} = \frac{\mathbf{f}_{j}}{{\left\lVert \mathbf{f}_{j} \right\rVert}_2} \odot \mathbf{\gamma}
\end{equation}
where ${\left\lVert \cdot \right\rVert}_2$ is the $L_2$ norm of a vector. 
The normalized feature vectors $\{ \hat{\mathbf{f}}_3 , \hat{\mathbf{f}}_4 , \hat{\mathbf{f}}_5 \}$ are concatenated into $\mathbf{x}_{c} \in \mathcal{R}^{d_{c}}$. 
To prevent over-fitting and improve generalization ability, a Dropout layer with dropout rate as $p_{d}$ is placed after the normalization. 
Feature fusion is done with a fully connected layer followed by Batch Normalization (BN).  
\begin{equation}
    \mathbf{x} = {\rm BN}({\rm Dropout}(\mathbf{x}_{c}) \mathbf{W} + \mathbf{b}), \ \mathbf{x} \in \mathcal{R}^{d}
\end{equation}
where $\mathbf{x}$ is the fused feature vector for image $\mathcal{I}$, and $d$ is the embedding dimension of $\mathbf{x}$.

\subsubsection{Context Aggregation via Transformer}
\begin{figure}[tb]
    \centering
    \includegraphics[page=3, width=0.96\columnwidth]{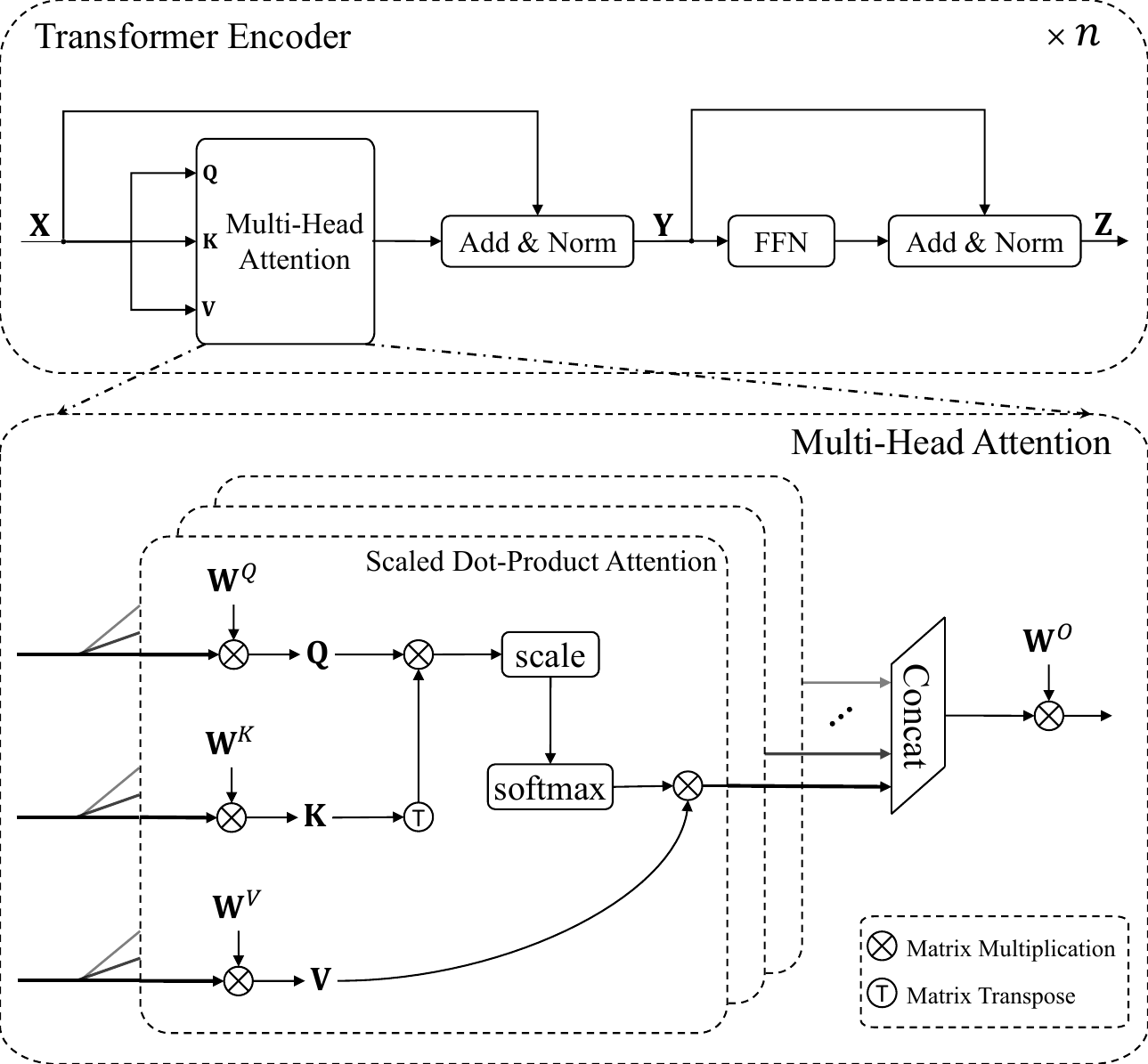}
    \caption{Architecture of Transformer encoder and Multi-Head Attention. 
            The Transformer encoder layer consists of a Multi-Head Attention layer and a feed-forward network. Stacking multiple encoder layers builds a Transformer encoder.
            The Multi-Head Attention concatenates and fuses the outputs from multiple scaled dot-product attention sub-modules. 
            }
    \label{fig:transformer encoder}
\end{figure}

The multi-block feature fusion module fuses low-level features in the shallower block with the high-level semantic features in the deeper block. However, feature vector $\mathbf{x}$ only captures the appearance of an independent image.
As we discussed before, persons or vehicles in the same identity but captured under different view angles provide complementary information to each other, which is known as context.
Together with its context, an image forms a comprehensive description for an object, which is more robust than a single image can achieve.
To aggregate the contextual cues in multiple images, we feed the sequence ($k_1$ nearest neighbors) into a Transformer encoder where Multi-Head Attention (MHA) updates each single image descriptor $\mathbf{x}$ with a weighted average of all the other image descriptors based on scaled dot-product similarity.
A feed-forward network (FFN) adds non-linearity to the aggregation.
The Transformer encoder is visualized in \cref{fig:transformer encoder}.

Let's denote the output sequence $\{\mathbf{x}_i\}_{i=1}^{k_1}$ from the multi-block feature fusion module as $\mathbf{X} \in \mathcal{R}^{k_1 \times d}$. 
The scaled dot-product attention first maps embeddings to Queries ($\mathbf{Q} \in \mathcal{R}^{k_1 \times d_s}$), Keys ($\mathbf{K} \in \mathcal{R}^{k_1 \times d_s}$) and Values ($\mathbf{V} \in \mathcal{R}^{k_1 \times d_s}$) with three learnable projection matrices. 
After that, the similarity between Queries and Keys aggregates Values together,
\begin{equation}
    {\rm MHA}(\mathbf{Q}, \mathbf{K}, \mathbf{V}) = softmax(\frac{\mathbf{Q} \mathbf{K}^{T}} {\sqrt{d_s}}) \mathbf{V}
\end{equation}
The multi-head structure refers to concatenating the outputs from multiple scaled dot-product attention modules and fusing them with a learnable projection $\mathbf{W}^O$.
It encapsulates complex relationships amongst different elements in the sequence by forcing each head to focus on some specific parts of the input embeddings.
The output is added to $\mathbf{X}$ and finally normalized via Layer Normalization (LN),
\begin{equation}
    \mathbf{Y} = {\rm LN}(\mathbf{X} + {\rm MHA}(\mathbf{Q}, \mathbf{K}, \mathbf{V}))
\end{equation}

Besides MHA, the Transformer encoder layer contains a position-wise feed-forward network for non-linearity. 
It consists of two linear transformations with a ReLU activation,
\begin{equation}
    {\rm FFN}(\mathbf{Y}) = {\rm ReLU}(\mathbf{Y} \mathbf{W}_{1} + \mathbf{b_1}) \mathbf{W}_{2} + \mathbf{b_2}
\end{equation}
The final output of one complete Transformer encoder layer is
\begin{equation}
    \mathbf{Z} = {\rm LN}(\mathbf{Y} + {\rm FFN}(\mathbf{Y}))    
\end{equation}
Following the original structure, we stack $n$ Transformer encoder layers for the context aggregation purpose.
The number of heads in MHA is controlled by a hyper-parameter $h$.
We name this encoder as BaseEncoder.

\subsubsection{Memory Initialization}
The BaseEncoder augments each feature embedding with a weighted average of all the other embeddings. 
However, we are still not able to distinguish the falsely retrieved samples.
To achieve our purpose, we propose to first transfer the probe's context into a memory cell, then reconstruct each retrieved sample with information fetched from memory, and finally predict their pairwise relationships.
The transfer of the probe's context in ACP is memory initialization.

\begin{figure}[tb]
    \centering
    \includegraphics[page=4, width=0.96\columnwidth]{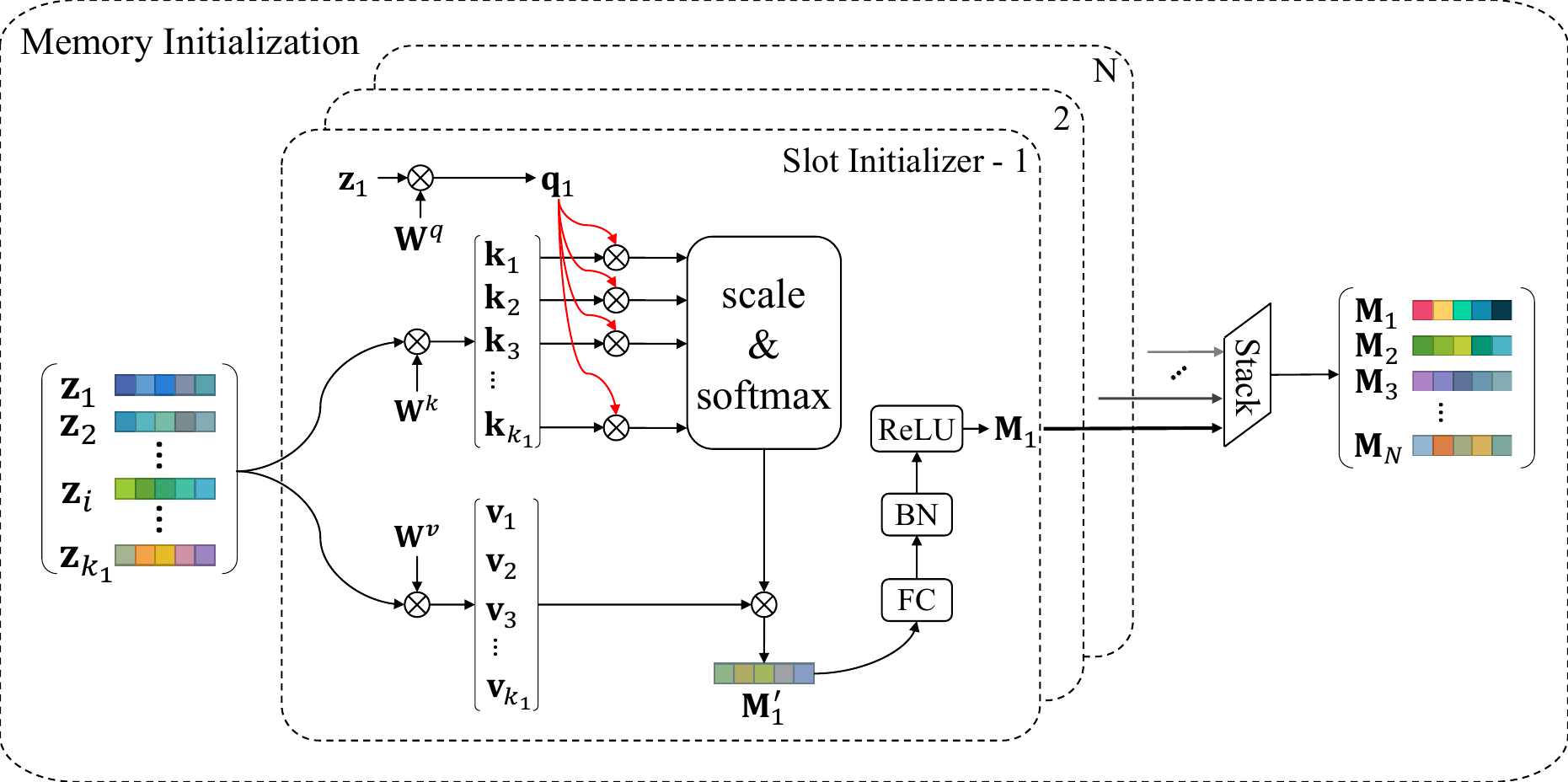}
    \caption{Memroy initialization. 
    $\{\mathbf{z}_i\}_{i=1}^{k_1}$ denotes the embeddings of $k_1$-nearest after BaseEncoder. 
    There are $N$ memory slots stacked to form a complete contextual memory cell.}
    \label{fig:memory initialization}
\end{figure}

The probe image itself is the only and most reliable information about the object we want to re-identify.
Thus, context can be obtained by comparing similarities between the probe and its corresponding $k_1$-nearest neighbors, i.e., $\mathbf{Z}$.
We store the aggregated contextual features in a Contextual Memory cell.
In other words, the memory collects different aspects of the probe image, like appearance in different angles, under different illuminations, etc. 
Therefore, the memory initialization mimics the summarization ability of humans who memorize the previously verified confident samples and utilize them for later identification.
The probe-related contextual feature aggregation, i.e., memory initialization, is realized with attention and illustrated in \cref{fig:memory initialization}.

\begin{figure}[tbp]
 \subfloat[]{
    \begin{minipage}[t]{0.96\columnwidth}
        \centering
        \includegraphics[width=\columnwidth]{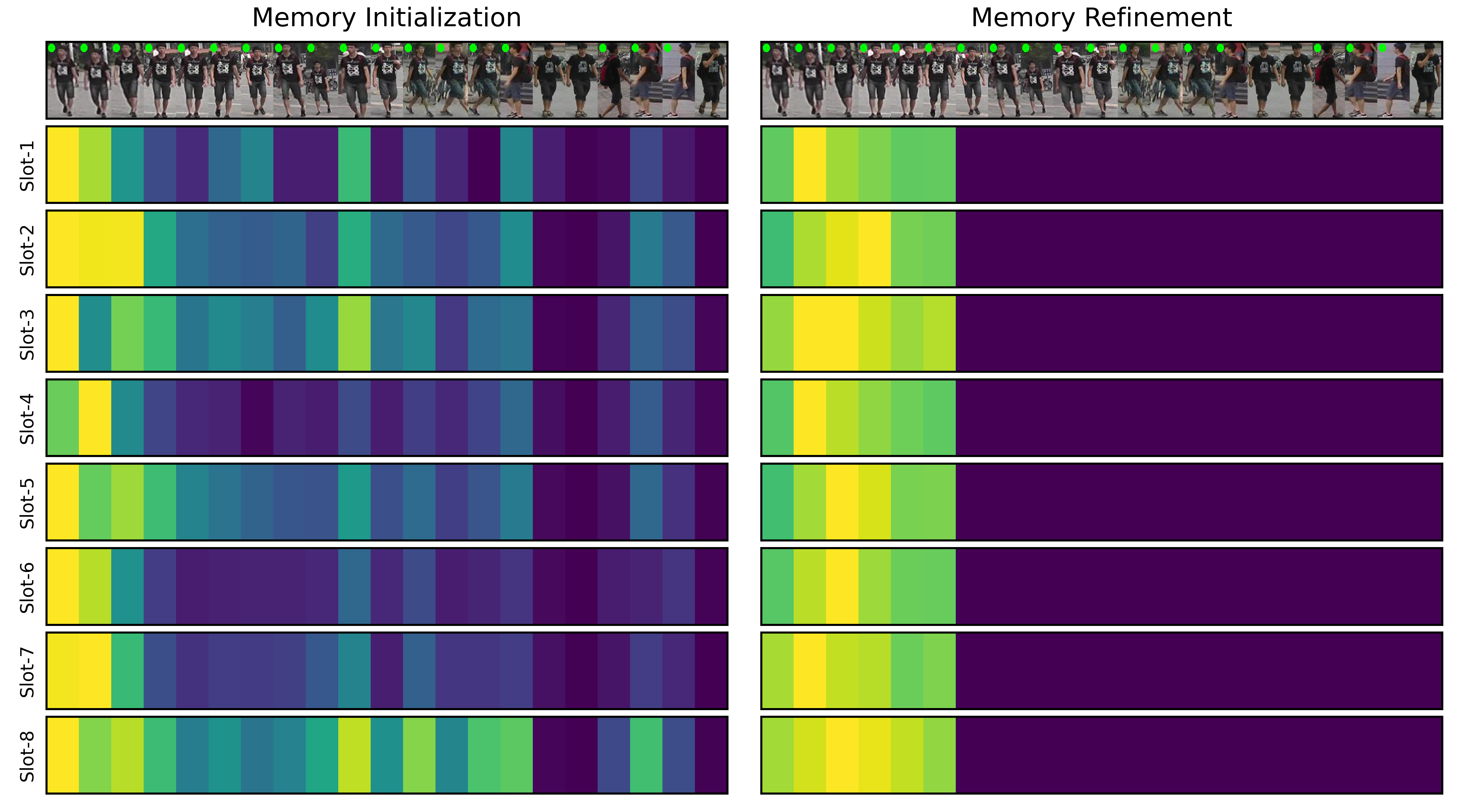}
    \end{minipage}
    \label{fig:visWeights-a}
 }
 \hspace{1px}
 \subfloat[]{
    \begin{minipage}[t]{0.96\columnwidth}
        \centering
        \includegraphics[width=\columnwidth]{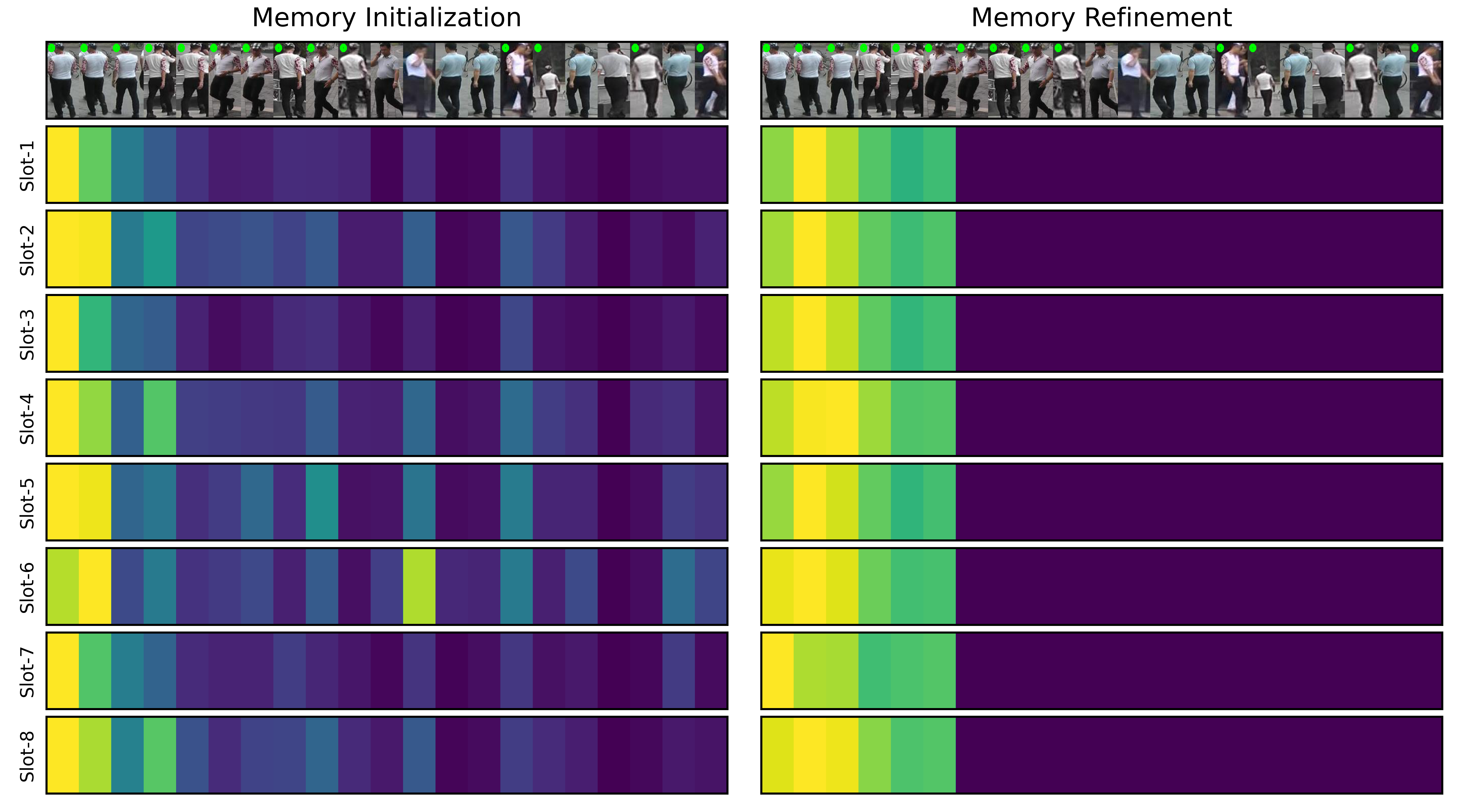}
    \end{minipage}
    \label{fig:visWeights-b}
 }
 \caption{Visualization of attention weights for memory initialization and refinement. 
 (a) and (b) are two different examples from Market1501.
 The image sequence denotes the query and top-20 gallery images retrieved by the baseline re-ID method.
 Slot-X indicates the X-th memory slot where X is an integer number. 
 We assign brighter colors for larger weights and vice versa.
 A green dot is added on the top-left corner for images in the same identity as the query.}
 \label{fig:visWeights}
\end{figure}

The attention first maps $\mathbf{Z}$ to Queries, Keys, and Values.
Instead of producing query vectors for all the elements, we only consider the probe embedding $\mathbf{z}_1$. 
This restriction guides our model to collect the probe-specific features from the local neighbors. 
Denote three transformation matrices as $\mathbf{W}^{q} \in \mathcal{R}^{d \times d_{m}}$, $\mathbf{W}^{k} \in \mathcal{R}^{d \times d_{m}}$ and $\mathbf{W}^{v} \in \mathcal{R}^{d \times d_{m}}$ where $d_{m}$ is the sub-feature dimension and $d_{m}$ is smaller than $d$.
The transformation can be formulated as,
\begin{equation}
    \begin{split}
        &\mathbf{q} = \mathbf{z}_{1} \mathbf{W}^{q} \\
        &\mathbf{K} = \mathbf{Z} \mathbf{W}^{k} \\
        &\mathbf{V} = \mathbf{Z} \mathbf{W}^{v} \\
    \end{split}
    \label{eq:qK similarity}
\end{equation}
We choose dot-product for similarity calculation and normalize them with softmax,
\begin{equation}
    \begin{split}
        \mathbf{M}_{i}^{'} &= softmax(\frac{\mathbf{q} \mathbf{K}^{T}} {\mu}) \mathbf{V}\\
                           &= softmax(\frac{\mathbf{z}_{1} \mathbf{W}^{q} {(\mathbf{Z} \mathbf{W}^{k})}^{T}} 
                              {\mu}) \mathbf{Z} \mathbf{W}^{v}
    \end{split}
    \label{equation:memory initialization}
\end{equation}
Here, $\mu$ is a learnable scale parameter that controls the non-linearity of softmax. 
A visualization of the combination weights is shown in \cref{fig:visWeights-a}.
\graphicspath{ {./images/}}

Finally, we insert a fully connected layer together with batch normalization and ReLU activation after attention.
Thus, a memory slot $\mathbf{M}_{i}$ is generated by,
\begin{equation}
    \mathbf{M}_{i} = {\rm ReLU}({\rm BN}(\mathbf{M}_{i}^{'} \mathbf{W} + \mathbf{b}))
\end{equation}
The fully connected layer expands $\mathbf{M}_{i}^{'}$ back to the original embedding dimension. 
As shown in \cref{fig:memory initialization}, the memory cell comprises $N$ independent memory slots.

\subsubsection{Memory Refinement}
The memory refinement scheme is proposed to eliminate the feature pollution from noisy false matches.
Different from other neural memory networks \cite{yuan2019memory, yang2019visual} whose memory writing or updating loops samples in a sequence one by one, we update $\mathbf{M}$ through the attention between the memory cell and top-ranked highly-confident samples parallelly. 

\begin{figure}[tb]
    \begin{minipage}[tb]{0.96\columnwidth}
        \centering
        \subfloat[Memroy refinement]
        {\includegraphics[page=5, width=0.96\columnwidth]{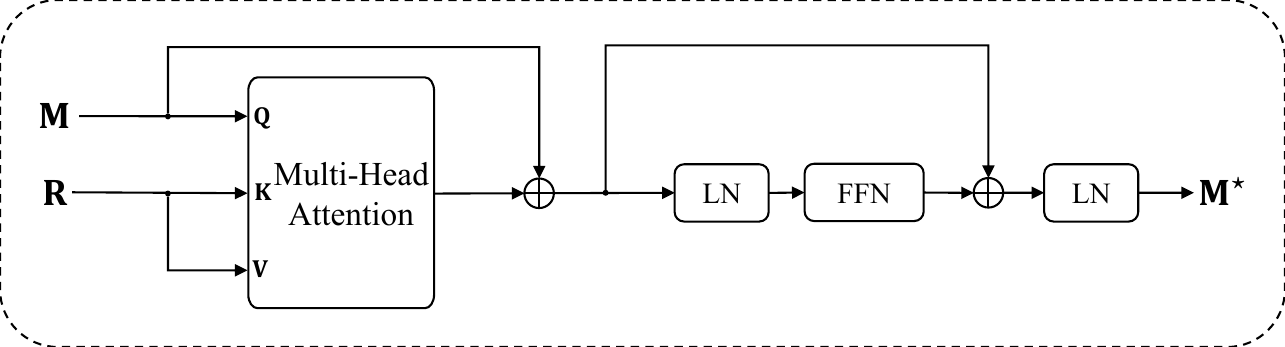}
        \label{fig:memory refinement}}
    \end{minipage}
    \vskip\baselineskip
    \begin{minipage}[tb]{0.96\columnwidth}
        \centering
        \subfloat[Feature reconstruction and Correlation Prediction]
        {\includegraphics[page=6, width=0.96\columnwidth]{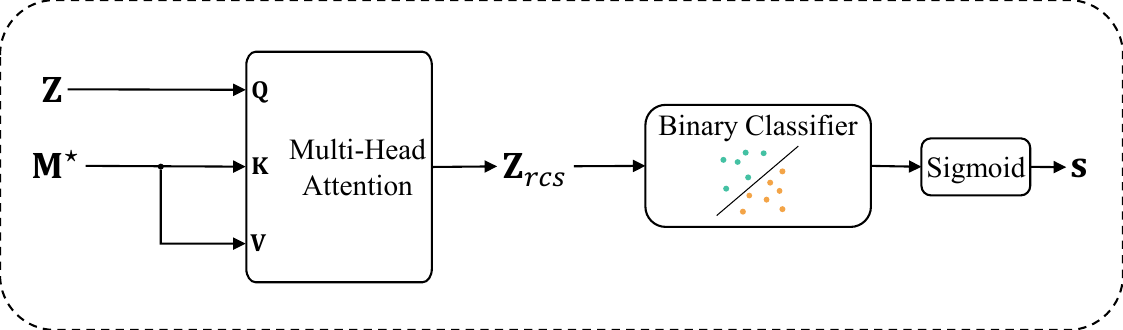}
        \label{fig:feature reconstruction}}
    \end{minipage}
    \caption{
            (a) Memory refinement updates the memory cell by aggregating relevant information from the refinement sequence.
            (b) Feature reconstruction reversely builds each feature embedding through a combination of multiple memory slots.
                The binary classifier separates the reconstructed embeddings predicting the correlations.
            }
\end{figure}

As mentioned above, $\mathbf{M}$ aggregates multiple aspects of a probe image from $\mathbf{Z}$. 
The key point of discriminability relies on the dot-product similarity between $\mathbf{q}$ and $\mathbf{K}$ in \cref{equation:memory initialization}.
Features of false matches might be transferred into $\mathbf{M}$ with inaccurate attention weights. 
As shown in \cref{fig:visWeights-b}, memory slot-6 assigns relatively large weights to some falsely retrieved samples during initialization, which will affect the subsequent feature reconstruction and correlation prediction. 
Hence, we add a memory refinement module after initialization illustrated in \cref{fig:memory refinement} that forces the memory to be updated on high-quality samples to prevent feature deviation. That is to say, the memory should `Remain True to the Original Aspiration'.

Given the BaseEncoder output sequence $\mathbf{Z}$, we chuck it into two parts with the upper part as $\mathbf{R} = \{\mathbf{z}_i\}_{i=1}^{k_2}$.
The sequence $\mathbf{R}$ is named as refinement sequence, which includes the first $k_2$ samples of $\mathbf{Z}$ ($k_2$ nearest neighbors of the probe).
Our memory refinement combines features through attention which converts $\mathbf{M}$ to Queries, and $\mathbf{R}$ to Keys and Values.
One thing worth mentioning is that the softmax here in MHA is applied on the $k_2$ dimension of $\mathbf{R}$. 
In other words, each memory slot refines itself by extracting information of interest from the refinement sequence. 
We visualize the combination weights in \cref{fig:visWeights-b}.
Memory initialization and refinement cooperate to exclude interfering features of falsely retrieved samples from entering the memory cell. 
We use this discriminative Contextual Memory $\mathbf{M}^{\star}$ for feature reconstruct and correlation prediction afterward.

\subsubsection{Feature Reconstruction and Correlation Prediction}
The goal of ACP is to accurately predict the correlations between an image and its local neighbors.
Since features stored in the memory cell are probe-specific, neighbors reconstructed from the memory will exhibit distinct patterns relevant to their identities. 
In other words, falsely retrieved samples after reconstruction will be different from those correct matches because they share less common features with the memory. 
A binary classifier can leverage the feature divergence to distinguish correct matches from strong distractors, predicting the correlations.

The implementation of feature reconstruction is shown in \cref{fig:feature reconstruction}.
The MHA compares the similarity between each neighbor and the Contextual Memory, which can be regarded as the reverse process of the memory initialization (see \cref{fig:memory initialization}).
Likewise, a learnable projection maps the $\mathbf{Z}$ to Queries, and the other two projections convert $\mathbf{M}^{\star}$ to Keys and Values.
Each feature embedding is reconstructed by finding a combination of multiple memory slots.
Let us denote the reconstruction output as $\mathbf{Z}_{rcs}$.
Unlike Transformer encoder layers where attention outputs are added to the input sequence as residuals, we keep the attention outputs as the final reconstruction results because we reconstruct features for correlation prediction rather than updating the input sequence.

Finally, a binary classifier separates the reconstructed embeddings. 
Denote the weight matrix of the classifier as $\mathbf{W}_{c} \in \mathcal{R}^{d \times 1}$, and the bias as $b_c$. 
The correlations prediction is,
\begin{equation}
    \mathbf{s} = Sigmoid(\mathbf{Z}_{rcs} \mathbf{W}_{c} + b_c), \ \mathcal{R}^{k_1 \times d} \to \mathcal{R}^{k_1}
\end{equation}
where $Sigmoid(\cdot)$ is the element-wise sigmoid activation. 
We choose Focal Loss \cite{lin2017focal} to supervise the network training. Denote $p_i$ as,
\begin{equation}
    p_i=\begin{cases}
        s_i   & \text{if} \quad y=1\\
        1-s_i & \text{otherwise}
        \end{cases}
\end{equation}
where $s_i$ is the i-th element of vector $\mathbf{s}$, and $y$ is the corresponding binary label (equals 1 if in the same identity as the probe). Therefore, the final loss is formulated as,
\begin{equation}
    FL(p_i) = \sum_{i=1}^{k_1}{-(1-p_i)^\gamma log(p_i)}
\end{equation}
The focusing parameter $\gamma$ balances weights between easy/hard examples.

\section{Experiments}
\subsection{Datasets and Evaluation metrics}
\begin{table}[]
    \caption{Re-ID datasets statistics. `IDs' denotes the number of different identities. `Imgs' represents the number of captured images. `Cams' refers to the number of unique cameras. `S', `M' and `L' indicate three different test set partitions namely small, medium and large for VERI-Wild and VehicleID.}
    \begin{adjustbox}{width=\columnwidth,center}
        \setlength\tabcolsep{1.5pt}
        \begin{tabular}{cc|c|c|c|c|c|c|c|c|c|c|}
            \cline{3-12}
                                                            &                      & \multicolumn{10}{c|}{ReID Datasets}                                                                                                                          \\ \hline
            \multicolumn{2}{|c|}{Splits}                                          & VeRi-776             & \multicolumn{3}{c|}{VERI-Wild} & \multicolumn{3}{c|}{VehicleID} & Market1501           & DukeMTMC             & MSMT17                \\ \hline
            \multicolumn{1}{|c|}{\multirow{3}{*}{Train}}   & IDs                  & 576                  & \multicolumn{3}{c|}{30671}     & \multicolumn{3}{c|}{13164}     & 751                  & 702                  & 1041                  \\ \cline{2-12} 
            \multicolumn{1}{|c|}{}                         & Imgs                 & 37746                & \multicolumn{3}{c|}{277797}    & \multicolumn{3}{c|}{113346}    & 12936                & 16522                & 30248                 \\ \cline{2-12} 
            \multicolumn{1}{|c|}{}                         & Cams                 & 20                   & \multicolumn{3}{c|}{173}       & \multicolumn{3}{c|}{/}         & 6                    & 8                    & 15                    \\ \hline
            \multicolumn{1}{|c|}{\multirow{4}{*}{Query}}   & \multirow{2}{*}{IDs} & \multirow{2}{*}{200} & S        & M        & L        & S        & M        & L        & \multirow{2}{*}{750} & \multirow{2}{*}{702} & \multirow{2}{*}{3060} \\ \cline{4-9}
            \multicolumn{1}{|c|}{}                         &                      &                      & 3000     & 5000     & 10000    & 800      & 1600     & 2400     &                      &                      &                       \\ \cline{2-12} 
            \multicolumn{1}{|c|}{}                         & Imgs                 & 1678                 & 3000     & 5000     & 10000    & 800      & 1600     & 2400     & 3368                 & 2228                 & 11659                 \\ \cline{2-12} 
            \multicolumn{1}{|c|}{}                         & Cams                 & 19                   & 105      & 113      & 126      & \multicolumn{3}{c|}{/}         & 6                    & 8                    & 15                    \\ \hline
            \multicolumn{1}{|c|}{\multirow{3}{*}{Gallery}} & IDs                  & 200                  & 3000     & 5000     & 10000    & 800      & 1600     & 2400     & 751                  & 1110                 & 3060                  \\ \cline{2-12} 
            \multicolumn{1}{|c|}{}                         & Imgs                 & 11579                & 38861    & 64389    & 128517   & 5693     & 11777    & 17377    & 15913                & 17661                & 82161                 \\ \cline{2-12} 
            \multicolumn{1}{|c|}{}                         & Cams                 & 19                   & 146      & 153      & 161      & \multicolumn{3}{c|}{/}         & 6                    & 8                    & 15                    \\ \hline
        \end{tabular}
    \end{adjustbox}
    \label{table:dataset}
    \end{table}

\subsubsection{Datasets}
We evaluate the proposed method on six widely-used re-ID datasets including three person re-ID datasets, i.e., Marker1501\cite{Market1501}, DukeMTMC-ReID\cite{DukeMTMC} and MSMT17\cite{MSMT17} as well as three vehicle re-ID datasets, i.e., VERI-Wild \cite{VeRiWild}, VehicleID \cite{VehicleID} and VeRi-776 \cite{VeRi}.
An overall statistical comparison on identities, images, and cameras of the datasets is in \cref{table:dataset}.
Especially, MSMT17 and VERI-Wild are closer to real-world scenarios because of the large number of images, cameras, and unique identities.
For example, VERI-Wild is collected from a CCTV surveillance system consisting of 174 cameras over one month under unconstrained conditions.
The MSMT17 contains both indoor and outdoor cameras operating at multiple time slots, resulting in severe lighting changes.

\subsubsection{Evaluation Metrics}
Following \cite{Market1501}, we choose two metrics to evaluate the re-ID performance. 
Cumulated Matching Characteristics (CMC) measures the probability that the ground-truth appears in the top-$k$ of the ranking list. 
Here, we report CMC@1 which is generally considered as a reflection of the ability to retrieve the easiest samples. 
Mean average precision (mAP) calculates the averaged area under the Precision-Recall curve of all the query images. 
It measures the ability to retrieve all related images.

\subsection{Implementation Details}
\subsubsection{Re-ID Baseline}
The baseline network is proposed by \cite{luo2019strong} which adopts a single-branch ResNet-50 as the backbone feature extractor.
In general, the strong baseline uses several following techniques to stabilize training and improve performance. 
For example, Learning Rate Warm-Up increases the learning rate linearly in the initial training stage to slow down over-fitting and maintain the stability of deeper layers. 
Batch Normalization (BN) Neck relieves the inconsistency between cross-entropy identity loss and triplet loss. 
Random Erasing Augmentation (REA) and Label Smoothing both aim at reducing over-fitting and encouraging robust features.
The implementation comes from FastReID\hyperfootnote{https://github.com/JDAI-CV/fast-reid}\cite{he2020fastreid} which is a powerful open-source toolbox designed for general instance re-ID.

\subsubsection{Attention-based Correlation Predictor}
To facilitate the usage of our method, we introduce two models based on the capacity of ACP, i.e., a basic model for regular-sized datasets and an XL version for large datasets. The specific model settings are listed in \cref{table:ACP}.
We choose ACP-Basic for datasets Market1501, DukeMTMC, VeRi-776, and MSMT17.
For VERI-Wild and VehicleID, we use ACP-XL.
We also calculate the total number of parameters and Multiply-Accumulate Operations (MACCs) for ACP-Basic and ACP-XL in \cref{table:param and MACCs}. Some famous networks are added for better comparison.
We can observe that even for ACP-XL that predicts top-$25$ samples, the MACCs of 1.26G is still lower than ResNet-18 with MACCs of 1.82G. 
The MACCs of ACP-Basic are smaller than ShuffleNet and MobileNet, which are almost negligible. 

\begin{table}[!t]
    \renewcommand\arraystretch{0.8}
    \caption{Model settings of ACP. Two models are named as ACP-Basic and ACP-XL based on the model capacity. The $\gamma$ belongs to Focal Loss.}
    \begin{adjustbox}{width=0.5\columnwidth,center}
        \setlength\tabcolsep{2pt}
        \begin{tabular}{@{}l|ccccccc@{}}
            \toprule
                                       & $n$                & $N$                & $h$                 & $d_m$                & $\gamma$           & $d$                   & $p_d$                \\ \midrule
            \multirow{2}{*}{ACP-Basic} & \multirow{2}{*}{3} & \multirow{2}{*}{8} & \multirow{2}{*}{16} & \multirow{2}{*}{32}  & \multirow{2}{*}{2} & \multirow{2}{*}{256}  & \multirow{2}{*}{0.4} \\
                                       &                    &                    &                     &                      &                    &                       &                      \\
            \multirow{2}{*}{ACP-XL}    & \multirow{2}{*}{3} & \multirow{2}{*}{8} & \multirow{2}{*}{16} & \multirow{2}{*}{128} & \multirow{2}{*}{1} & \multirow{2}{*}{1024} & \multirow{2}{*}{0}   \\
                                       &                    &                    &                     &                      &                    &                       &                      \\ \bottomrule
            \end{tabular}
    \end{adjustbox}
    \label{table:ACP}
\end{table}
\begin{table*}[htb]
    \caption{Comparasion of total parameters and Multiply-Accumulate Operations (MACCs) between different models.}
    \begin{adjustbox}{width=\textwidth,center}
        \begin{tabular}{ccccccccc}
            \hline
                                                                                             & ACP-Basic       & ACP-XL          & AlexNet \cite{krizhevsky2012imagenet}          & VGG16 \cite{simonyan2014very}             & ResNet-18 \cite{he2016deep}        & ResNet-50 \cite{he2016deep}        & ShuffleNet-V2 \cite{ma2018shufflenet}    & Mobilenet-V2 \cite{sandler2018mobilenetv2}     \\ \hline
            \multicolumn{1}{c|}{\begin{tabular}[c]{@{}c@{}}Num of\\ Params (M)\end{tabular}} & 4.61            & 62.48           & 61.10                 & 138.36                 & 11.69                 & 25.56                 & 2.28                  & 3.50                  \\ \hline
            \multicolumn{1}{c|}{\multirow{2}{*}{MACC (G)}}                                   & 0.10 ($k_1=25$) & 1.26 ($k_1=25$) & \multirow{2}{*}{0.77} & \multirow{2}{*}{15.61} & \multirow{2}{*}{1.82} & \multirow{2}{*}{4.14} & \multirow{2}{*}{0.15} & \multirow{2}{*}{0.33} \\
            \multicolumn{1}{c|}{}                                                            & 0.19 ($k_1=50$) & 2.40 ($k_1=50$) &                       &                        &                       &                       &                       &                       \\ \hline
            \end{tabular}
    \end{adjustbox}
    \label{table:param and MACCs}
\end{table*}

To construct a training sequence, we find the $K=1000$ nearest neighbors for each image.
Suppose there are $m$ correct matches and $K-m$ false matches.
We randomly select $l_1$ samples from the union set of the correct matches and the first several false matches.
The refinement sequence length is set to $l_2$.
Multiple sequences are stacked together to form a training batch. We set the batch size as 16 for all the experiments.
The model training adopts Adam optimizer with weight decay as $5 \times 10^{-4}$. 
The initial learning rate is set to $2 \times 10^{-4}$.
Similar to \cite{luo2019strong}, we linearly warm up the learning process for the first 10 epochs with the warm-up factor equaling $0.1$. 
The parameter $\gamma$ in Focal loss balances easy and hard samples.
The testing parameter $k_1$ is determined empirically based on the average number of images per identity in gallery sets as suggested in \cite{zhang2020understanding}. Suppose there are $N$ images and $C$ identities, then $k_1 = N / C$.
All the experiments are conducted on a platform with 256GB RAM and 2 $\times$ Intel Xeon Silver 4214R CPU @ 2.40GHz. 
The GPU we use is a single GeForce RTX 2080Ti with 11GB VRAM.

\subsection{Performance Comparison}
In this subsection, we compare the performance and time usage of our re-ranking approach with several widely-used re-ranking methods on the datasets mentioned above. 
Results are reported and analyzed in \cref{sec:Results with Selected Parameters}. 
We choose parameters based on the recommendations tested in their original papers. 
Because re-ranking methods are generally sensitive to parameter selection, we also studied the influence of hyper-parameters in \cref{sec:Parameter Sensitivity}.

\begin{table}[!t]
    \caption{The detailed testing hyper-parameters of all the competing methods. $N$ and $C$ refer to the number of images and identities in a gallery set, respectively.}
    \begin{adjustbox}{width=\columnwidth,center}
\begin{tabular}{@{}c|c|cc|cccc|ccc@{}}
\toprule
\multirow{2}{*}{} & AQE                            & \multicolumn{2}{c|}{$\alpha$QE}                     & $k$-reciprocal                 & SCA               & GNN                & Ours               & \multicolumn{3}{c}{ECN}                                       \\ \cmidrule(l){2-11} 
                  & $k$                            & $k$                            & $\alpha$           & \multicolumn{2}{c}{$k_1$}                          & \multicolumn{2}{c|}{$k_2$}              & $t$                & $q$                & $K$                 \\ \midrule
MSMT17            & \multirow{6}{*}{$\frac{N}{C}$} & \multirow{6}{*}{$\frac{N}{C}$} & \multirow{6}{*}{3} & \multicolumn{2}{c}{\multirow{6}{*}{$\frac{N}{C}$}} & \multicolumn{2}{c|}{\multirow{4}{*}{6}} & \multirow{5}{*}{3} & \multirow{5}{*}{8} & \multirow{5}{*}{25} \\
Market1501        &                                &                                &                    & \multicolumn{2}{c}{}                               & \multicolumn{2}{c|}{}                   &                    &                    &                     \\
DukeMTMC          &                                &                                &                    & \multicolumn{2}{c}{}                               & \multicolumn{2}{c|}{}                   &                    &                    &                     \\
VERI-Wild         &                                &                                &                    & \multicolumn{2}{c}{}                               & \multicolumn{2}{c|}{}                   &                    &                    &                     \\
VehicleID         &                                &                                &                    & \multicolumn{2}{c}{}                               & \multicolumn{2}{c|}{4}                  &                    &                    &                     \\
VeRi-776          &                                &                                &                    & \multicolumn{2}{c}{}                               & \multicolumn{2}{c|}{10}                 & 5                  & 12                 & 60                  \\ \bottomrule
\end{tabular}
    \end{adjustbox}
    \label{table:testing_parameter}
\end{table}

\begin{table*}[]
    \centering
    \caption{Performance (\%) comparison. The best and second-best results are makred in red and blue, respectively. `S', `M' and `L' denotes `Small', `Medium' and `Large', respectively. For time usage, all the methods will be run on GPU whenever available, and $\star$ indicates the success of performing GPU re-ranking. Otherwise, algorithms will be tested on CPU. For re-ranking methods that fails to run, an `-' is used to fill the blank.}
   
        \setlength\tabcolsep{3pt}
        \begin{tabular}{@{}c|c|cccccccccc@{}}
            \toprule
            \multicolumn{2}{c|}{}                             &       & Baseline                     & AQE                          & $\alpha$QE                   & $k$-reciprocal               & \begin{tabular}[c]{@{}c@{}}$\alpha$QE +\\ $k$-reciprocal\end{tabular} & SCA                          & GNN                          & ECN                          & Ours                         \\ \midrule
            \multicolumn{2}{c|}{}                             & CMC@1 & 73.75                        & 76.82                        & 79.04                        & {\color[HTML]{333333} 78.92} & 78.83                                                                 & {\color[HTML]{3166FF} 79.10} & -                            & -                            & {\color[HTML]{FE0000} 81.81} \\
            \multicolumn{2}{c|}{}                             & mAP   & 50.31                        & 64.26                        & 65.42                        & 67.12                        & {\color[HTML]{3166FF} 69.94}                                          & 69.08                        & -                            & -                            & {\color[HTML]{FE0000} 71.99} \\
            \multicolumn{2}{c|}{\multirow{-3}{*}{MSMT17}}     & time  & -                            & 45.9s                        & 50.1s                        & 430.6s                       & -                                                                     & 1339.1s                      & -                            & -                            & 3.6s$^\star$                 \\ \midrule
            \multicolumn{2}{c|}{}                             & CMC@1 & 96.07                        & 96.01                        & {\color[HTML]{FE0000} 97.44} & 96.54                        & 96.96                                                                 & 96.60                        & 96.84                        & {\color[HTML]{333333} 97.14} & {\color[HTML]{3166FF} 97.20} \\
            \multicolumn{2}{c|}{}                             & mAP   & 78.56                        & 84.54                        & 85.95                        & 85.61                        & {\color[HTML]{FE0000} 87.58}                                          & 84.52                        & 86.10                        & 85.52                        & {\color[HTML]{3166FF} 87.26} \\
            \multicolumn{2}{c|}{\multirow{-3}{*}{VeRi-776}}   & time  & -                            & 2.2s                         & 2.2s                         & 39.1s                        & -                                                                     & 44.3s                        & 1.8ms$^\star$                & 13.0s                        & 1.7s$^\star$                 \\ \midrule
                                        &                     & CMC@1 & 93.27                        & 85.41                        & 92.67                        & {\color[HTML]{3166FF} 93.84} & 91.83                                                                 & 92.77                        & -                            & 90.63                        & {\color[HTML]{FE0000} 95.18} \\
                                        &                     & mAP   & 76.74                        & 71.64                        & 79.37                        & {\color[HTML]{3166FF} 80.32} & 78.94                                                                 & 79.65                        & -                            & 79.68                        & {\color[HTML]{FE0000} 88.16} \\
                                        & \multirow{-3}{*}{S} & time  & -                            & 2.4s                         & 3.6s                         & 111.4s                       & -                                                                     & 211.4s                       & -                            & 54.2s                        & 11.8s$^\star$                \\ \cmidrule(l){2-12} 
                                        &                     & CMC@1 & 90.04                        & 81.04                        & 89.50                        & {\color[HTML]{3166FF} 90.66} & 88.01                                                                 & 89.52                        & -                            & -                            & {\color[HTML]{FE0000} 92.45} \\
                                        &                     & mAP   & 70.61                        & 65.51                        & 72.75                        & {\color[HTML]{3166FF} 73.94} & 72.18                                                                 & 73.30                        & -                            & -                            & {\color[HTML]{FE0000} 83.60} \\
                                        & \multirow{-3}{*}{M} & time  & -                            & 30.4s                        & 32.6s                        & 272.7s                       & -                                                                     & 701.8s                       & -                            & -                            & 18.9s$^\star$                \\ \cmidrule(l){2-12} 
                                        &                     & CMC@1 & 86.52                        & 76.20                        & 85.59                        & {\color[HTML]{3166FF} 87.09} & 82.80                                                                 & -                            & -                            & -                            & {\color[HTML]{FE0000} 89.64} \\
                                        &                     & mAP   & 62.56                        & 56.64                        & 63.92                        & {\color[HTML]{3166FF} 65.65} & 63.15                                                                 & -                            & -                            & -                            & {\color[HTML]{FE0000} 75.80} \\
            \multirow{-9}{*}{VERI-Wild} & \multirow{-3}{*}{L} & time  & -                            & 112.9s                       & 120.4s                       & 1221.5s                      & -                                                                     & -                            & -                            & -                            & 37.5s$^\star$                \\ \midrule
                                        &                     & CMC@1 & 80.45                        & {\color[HTML]{333333} 71.78} & {\color[HTML]{3166FF} 81.51} & 80.64                        & 80.38                                                                 & 79.71                        & 80.26                        & 75.94                        & {\color[HTML]{FE0000} 87.28} \\
                                        &                     & CMC@5 & 94.44                        & 92.18                        & {\color[HTML]{3166FF} 95.72} & 94.61                        & 94.99                                                                 & 91.80                        & 92.54                        & 91.52                        & {\color[HTML]{FE0000} 96.71} \\
                                        & \multirow{-3}{*}{S} & time  & -                            & 0.4ms$^\star$                & 7.5ms$^\star$                & 9.1s                         & -                                                                     & 7.3s                         & 1.3ms$^\star$                & 1.3s                         & 1.9s$^\star$                 \\ \cmidrule(l){2-12} 
                                        &                     & CMC@1 & {\color[HTML]{3531FF} 77.48} & 68.02                        & {\color[HTML]{333333} 77.35} & 77.22                        & 75.98                                                                 & 75.80                        & 76.26                        & 72.71                        & {\color[HTML]{FE0000} 82.58} \\
                                        &                     & CMC@5 & 91.28                        & 88.72                        & {\color[HTML]{3166FF} 91.86} & 90.71                        & 91.14                                                                 & 87.85                        & 88.75                        & 87.84                        & {\color[HTML]{FE0000} 93.83} \\
                                        & \multirow{-3}{*}{M} & time  & -                            & 0.5ms$^\star$                & 9.4ms$^\star$                & 21.3s                        & -                                                                     & 27.8s                        & 1.7ms$^\star$                & 6.8s                         & 3.9s$^\star$                 \\ \cmidrule(l){2-12} 
                                        &                     & CMC@1 & 75.33                        & 66.63                        & {\color[HTML]{3166FF} 76.17} & 75.65                        & 75.21                                                                 & 74.63                        & 75.21                        & 71.42                        & {\color[HTML]{FE0000} 81.72} \\
                                        &                     & CMC@5 & 88.79                        & 86.56                        & {\color[HTML]{3166FF} 89.66} & 88.21                        & 88.79                                                                 & 86.31                        & 87.36                        & 85.90                        & {\color[HTML]{FE0000} 91.87} \\
            \multirow{-9}{*}{VehicleID} & \multirow{-3}{*}{L} & time  & -                            & 0.4ms$^\star$                & 15.9ms$^\star$               & 36.4s                        & -                                                                     & 57.4s                        & 2.5ms$^\star$                & 14.9s                        & 5.3s$^\star$                 \\ \midrule
            \multicolumn{2}{c|}{}                             & CMC@1 & {\color[HTML]{333333} 93.85} & 94.24                        & 95.37                        & {\color[HTML]{333333} 95.34} & 95.37                                                                 & 95.19                        & 95.34                        & {\color[HTML]{FE0000} 95.61} & {\color[HTML]{3166FF} 95.46} \\
            \multicolumn{2}{c|}{}                             & mAP   & 86.31                        & 91.85                        & 93.08                        & 93.98                        & 93.97                                                                 & 94.14                        & {\color[HTML]{FE0000} 94.55} & {\color[HTML]{3166FF} 94.47} & 93.82                        \\
            \multicolumn{2}{c|}{\multirow{-3}{*}{Market1501}} & time  & -                            & 0.4ms$^\star$                & 6.8ms$^\star$                & 33.9s                        & -                                                                     & 59.9s                        & 3.2ms$^\star$                & 14.9s                        & 1.1s$^\star$                 \\ \midrule
            \multicolumn{2}{c|}{}                             & CMC@1 & 86.36                        & 89.95                        & {\color[HTML]{333333} 90.22} & 90.53                        & 90.26                                                                 & 90.04                        & 90.71                        & {\color[HTML]{FE0000} 91.11} & {\color[HTML]{3166FF} 90.84} \\
            \multicolumn{2}{c|}{}                             & mAP   & 76.98                        & 86.56                        & 86.92                        & 88.93                        & 89.15                                                                 & 89.34                        & {\color[HTML]{FE0000} 90.03} & {\color[HTML]{3166FF} 89.71} & 89.03                        \\
            \multicolumn{2}{c|}{\multirow{-3}{*}{DukeMTMC}}   & time  & -                            & 0.4ms$^\star$                & 11.9ms$^\star$               & 35.9s                        & -                                                                     & 62.7s                        & 4.1ms$^\star$                & 11.5s                        & 1.2s$^\star$                 \\ \bottomrule
            \end{tabular}
   
    \label{table:performance_time}
\end{table*}

\subsubsection{Comparisons with State-of-the-art Methods} \label{sec:Results with Selected Parameters}
We tested AQE \cite{chum2007total}, $\alpha$QE \cite{radenovic2018fine}, 
$k$-reciprocal \cite{zhong2017re}, SCA \cite{bai2016sparse}, GNN \cite{zhang2020understanding} and ECN \cite{sarfraz2018pose}.
The detailed testing parameters for all the competing methods are available in \cref{table:testing_parameter}. Based on their default settings, we estimate the optimal values for other datasets not evaluated in their papers. For AQE, $\alpha$QE, k-reciprocal, SCA, GNN, and ours, we set parameter controlling the number of first-order nearest neighbors equal to $N/C$ for fair comparison (the same amount of contextual information). The $k_2$ is estimated as a much smaller value than $k_1$ as suggested in \cite{zhong2017re} and \cite{zhang2020understanding} to avoid bringing in noisy samples. ECN is slightly different because it uses $t$ and $q$ to represent the first and second-order nearest neighbors, respectively. As recommended in \cite{sarfraz2018pose}, ECN should use strongest top neighbors in the first-order (small $t$) and expand these to few more at the second-order (relatively larger $q$). We follow this rule during its parameter selection for different datasets.
Experimental results for each method are listed in \cref{table:performance_time}.
The row named `$\alpha$QE + $k$-reciprocal' refers to perform $k$-reciprocal re-ranking after $\alpha$QE. 
For GNN, ECN and SCA re-ranking, some results are not available because of insufficient memory. 
Note that the implementation of AQE and $\alpha$QE here expand all the feature embeddings by considering the query and gallery sets as a whole.

\textbf{MSMT17} is the largest person re-ID dataset in our experiments, especially for the test set, which adds up to almost 100k images in total.
As shown in \cref{table:performance_time}, we ranked first for both CMC@1 and mAP.
The baseline CMC@1 improved from 73.75\% to 81.81\%, and baseline mAP increased to 71.99\% from 50.31\% with 21.68\% improvement.
The improvement proves that our method is better at handling complex relationships between the $k$-nearest neighbors.
Besides, we only require the access of first-order nearest neighbors for each probe embedding.

For time usage, our approach only takes 3.6s, which is 119.6 times faster than k-reciprocal re-ranking (430.6s).
Although k-reciprocal seems to have less calculation at first glance, it requires the calculation of k-reciprocal feature vectors, which indicate whether a gallery image exists in the k-reciprocal neighbor set of a query. 
This operation cannot be easily accelerated on GPU with large-scale parallelization since it is not standard matrix computations.
GNN re-ranking implements a CUDA kernel that speeds up the calculation of k-reciprocal features on GPU. 
However, finding k-reciprocal neighbors has to build an adjacency matrix in GPU VRAM whose space complexity is $O(N^2)$ where N is the total number of images in query and gallery sets. 
For dataset MSMT17 with 93,820 images, the adjacency matrix of 32-bit precision float number would take 32.79GB VRAM, which is too large to fit in commonly seen GPUs, not even considering the additional memory required by getting adjacency matrix's transpose and other calculations. 
ECN re-ranking is killed halfway by the system Out-Of-Memory killer after occupying 256GB RAM plus 60GB virtual memory.

\textbf{VeRi-776} has relative more images per identity on average. 
Therefore, we increase the neighborhood size $k_1$ accordingly.
Results show that we produce the second-best mAP, which equals 87.26\%.
Although our method does not rank first for CMC@1, there is only 0.24\% difference compared to the best result obtained by $\alpha$QE.
Our method achieved the balance between CMC@1 and mAP.

\textbf{VERI-Wild.}
We can significantly boost the baseline performance on the small, medium, and large subsets of VERI-Wild.
By comparing the results across different subsets, we can observe that the $k$-reciprocal rule begins to fail as the testing set becomes larger. 
The mAP improvements for $k$-reciprocal re-ranking are 3.58\%, 3.33\%, 3.09\% for small, medium, and large, respectively.
However, our method is robust to the test set size achieving 11.42\%, 12.99\%, and 13.24\%.
Some methods run out of memory halfway.

\textbf{VehicleID.}
The metric mAP is not provided because there is only one correct match in the gallery set for each query image. 
Instead, we report CMC@5, which measures the probability of having the correct match in top-$5$ candidates.
To avoid randomness, we test 10 times and record the average score. 
As shown in the table, our method outperforms other re-ranking approaches for all three subsets by a large margin.
Improvement of $k$-reciprocal re-ranking is limited, which might be caused by the insufficient contextual information in the gallery set (one image per identity).
The $\alpha$QE gains a slight improvement over the baseline by taking the query-to-query similarities into account.
Comparison between $\alpha$QE and our approach demonstrates the superiority of discriminative correlation prediction.

\textbf{Market1501.} 
Our method achieves the second-best CMC@1, which equals 95.46\% with only 0.15\% difference compared to the best obtained by ECN re-ranking.
For mAP, the best result 94.55\% is produced by GNN re-ranking, which is 0.73\% higher than ours. 
We ascribe this unsatisfying result to the lack of training data. 
As shown in \cref{table:dataset}, Marker1501 is the smallest amongst all six datasets with only 13k images for training.
The backbone itself has already been facing over-fitting problems, not even for building a re-ranking network without extra training data.

\textbf{DukeMTMC.} 
Similar to Market1501, our approach ranked second for CMC@1. 
We obtain 90.84\% which is 0.27\% lower than the best 91.11\%.
For mAP, all the methods bring considerable improvements, and our approach is 12.05\% higher than the baseline. 
The highest mAP is produced by ECN re-ranking. 

Our method obtains relatively inferior results on Market1501 and DukeMTMC, which is mainly caused by the lack of training data. 
Among the competing re-ranking approaches, only ACP requires training, whereas other algorithms adopt hand-designed rules.
During ACP's training, the baseline ResNet-50 is frozen to reduce computation costs.
Therefore, if the baseline has already perfectly classified all the training samples, no new knowledge is left for ACP to learn.
To verify this assumption, we tested the baseline's re-ID performance on each training set as shown in \cref{table:train_performance}.
We notice the baseline obtains almost 100\% mAP and CMC@1 on small datasets like Market1501, DukeMTMC and VeRi-776, indicating the severity of over-fitting. That is, it is hard for our feature expansion-based ACP to learn to move feature embeddings toward centers by training on well-classified samples.
By contrast, VehicleID and VERI-Wild are free from this issue thanks to the massive training data available.

\begin{figure}[tbp]
    \centering
    \subfloat[Market1501]{
        \includegraphics[width=0.46\columnwidth]{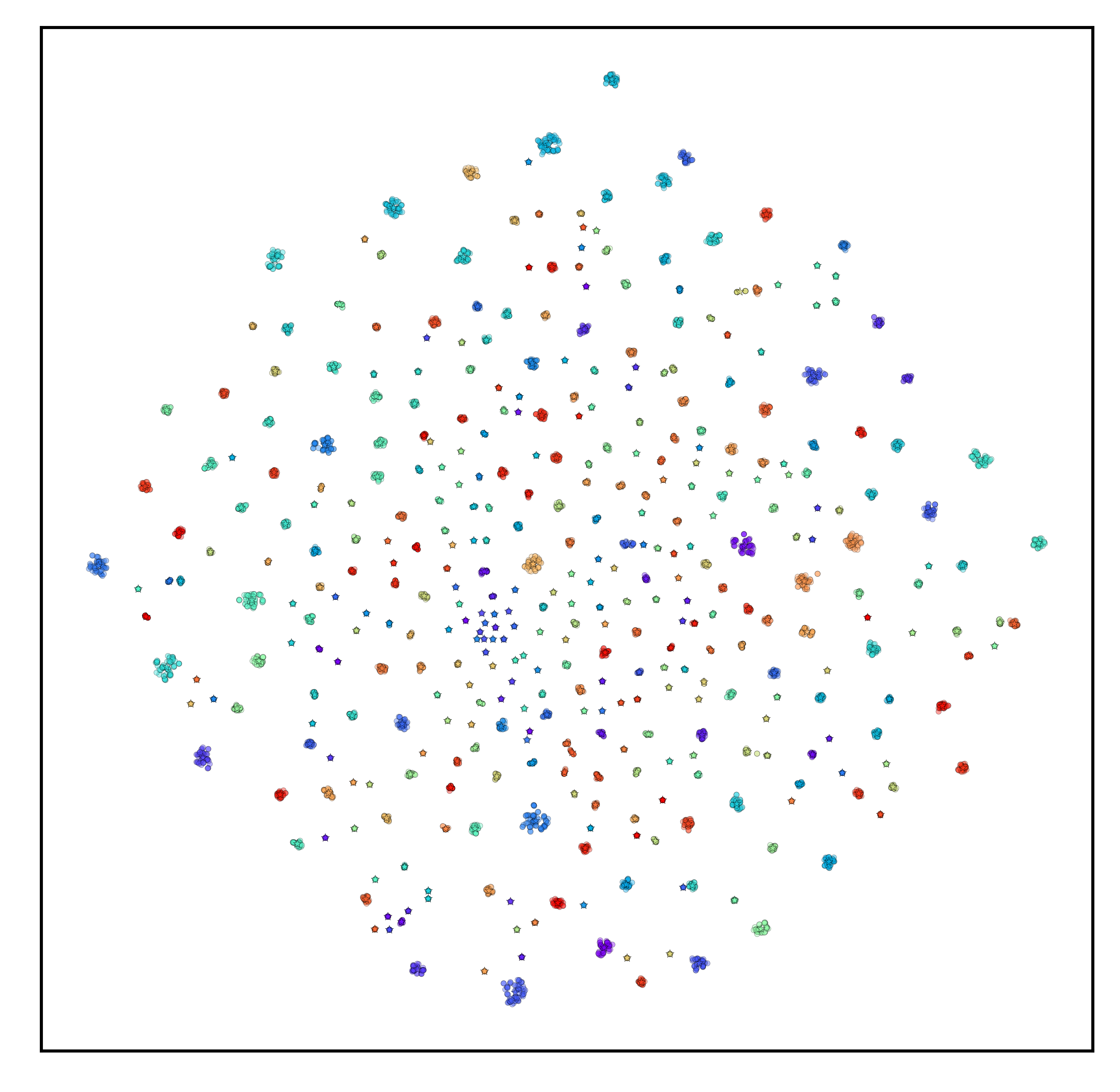}
    }
    \subfloat[VeRiWild]{
        \includegraphics[width=0.46\columnwidth]{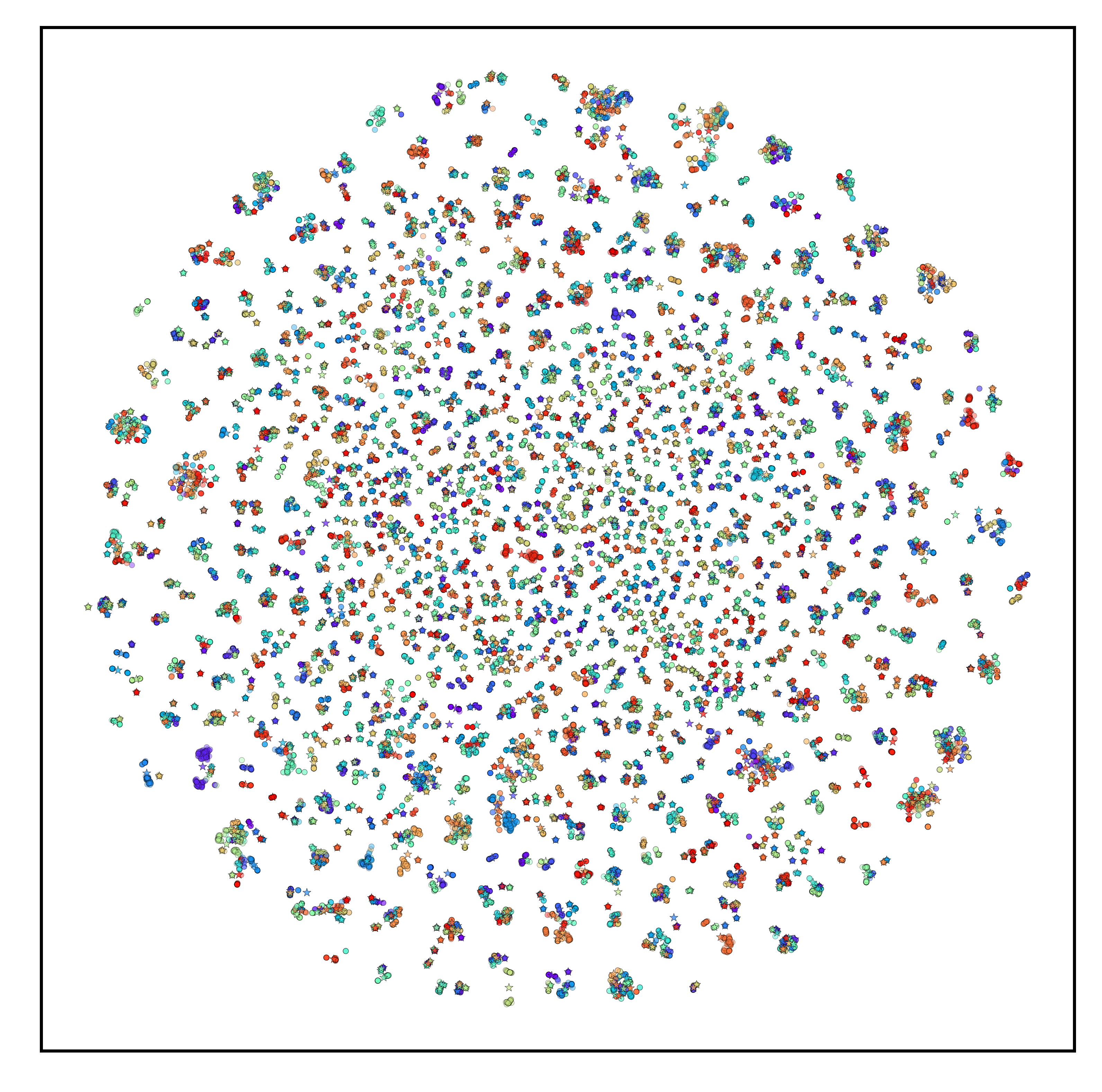}
    }
    \caption{Visualization of the feature embedding distribution after t-SNE dimension reduction. Different identities are shown in colors. The center location of each identity is marked with a star.
    }
    \label{fig:BoT_overfitting}
\end{figure}

To give an intuitive perception of each dataset's status, we visualize distributions of the extracted feature vectors for training sets using t-SNE \cite{van2008visualizing} in \cref{fig:BoT_overfitting}. 
For Market1501, feature embeddings in the same identity form tight groups, and different groups can be well separated.
This implies that the baseline network is able to distinguish all the persons and map each person's images to a unique location.
Because the baseline feature embeddings already show distinct differences if they are not in the same identity, training an ACP with satisfying generalization ability is difficult.
For VERI-Wild, we randomly chose 5,000 identities for visualization because a large amount of training samples makes the scatter plot too messy to analyze.
Even with significantly reduced samples, many identities still mix together, i.e., underfitting.
By constructing a proper training sequence, the proposed ACP learns to distinguish those miss classified samples with attention and memory.

Same as AQE and $\alpha$QE, ACP is a feature expansion-based method, which exploits the similarities between the baseline's embeddings. 
This kind of algorithm does not require obtaining the structural information of each sample though it is helpful in discriminating extremely similar false matches.
In summary, our ACP surpasses its pioneers AQE and $\alpha$QE comprehensively and can compete with neighborhood structure-based algorithms (e.g., k-reciprocal, SCA, GNN re-ranking) even when the training data is insufficient.

\begin{table}[tbp]
    \centering
    \caption{The performance of baseline ResNet-50 on different training sets.}
    \begin{adjustbox}{width=\columnwidth,center}
        \begin{tabular}{@{}c|cccccc@{}}
            \toprule
                   & Market1501 & DukeMTMC & MSMT17  & VeRi-776 & VehicleID & VERI-Wild \\ \midrule
            CMC@1  & 99.86\%    & 99.55\%  & 99.16\% & 99.72\%  & 93.62\%   & 81.53\%   \\
            CMC@5  & 99.97\%    & 99.98\%  & 99.92\% & 99.96\%  & 96.41\%   & 90.60\%   \\
            CMC@10 & 100.00\%   & 99.99\%  & 99.96\% & 100.00\% & 97.46\%   & 93.31\%   \\
            mAP    & 99.88\%    & 99.70\%  & 99.13\% & 99.21\%  & 74.98\%   & 59.64\%   \\ \bottomrule
            \end{tabular}
    \end{adjustbox}
    \label{table:train_performance}
\end{table}

\subsubsection{Parameter Sensitivity} \label{sec:Parameter Sensitivity}
Here, we study the influence of the testing parameters $k_1$ and $k_2$.

\begin{figure}[tb]
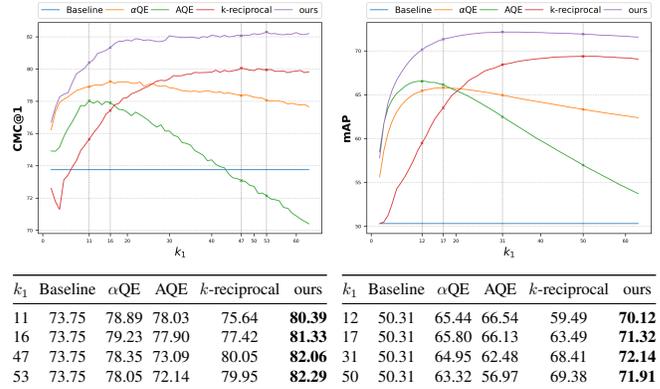

    \centering
    \subfloat[CMC@1 (\%) wrt. $k_1$]{
    \begin{minipage}[tb]{0.48\columnwidth}
        \begin{minipage}[tb]{\columnwidth}
            \centering
            \includegraphics[width=\columnwidth]{CMC1.png}
        \end{minipage}
        \vskip 0.15cm
        \begin{minipage}[tb]{\columnwidth}
            \centering
            \begin{adjustbox}{width=\columnwidth,center}
                \setlength\tabcolsep{3pt}
                \begin{tabular}{@{}cccccc@{}}
                    \toprule
                    $k_1$ & Baseline & $\alpha$QE & AQE     & $k$-reciprocal & ours             \\ \midrule
                    11    & 73.75  & 78.89    & 78.03 & 75.64      & \textbf{80.39} \\
                    16    & 73.75  & 79.23    & 77.90 & 77.42      & \textbf{81.33} \\
                    47    & 73.75  & 78.35    & 73.09 & 80.05      & \textbf{82.06} \\
                    53    & 73.75  & 78.05    & 72.14 & 79.95      & \textbf{82.29} \\ \bottomrule
                    \end{tabular}
            \end{adjustbox}
        \end{minipage}
    \end{minipage}}
    \subfloat[mAP (\%) wrt. $k_1$]{
    \begin{minipage}[tb]{0.48\columnwidth}
        \begin{minipage}[tb]{\columnwidth}
            \centering
            \includegraphics[width=\columnwidth]{mAP.png}
        \end{minipage}
        \vskip 0.15cm
        \begin{minipage}[tb]{\columnwidth}
            \centering
            \begin{adjustbox}{width=\columnwidth,center}
                \setlength\tabcolsep{3pt}
                \begin{tabular}{@{}cccccc@{}}
                    \toprule
                    $k_1$ & Baseline & $\alpha$QE & AQE     & $k$-reciprocal & ours             \\ \midrule
                    12    & 50.31  & 65.44    & 66.54 & 59.49      & \textbf{70.12} \\
                    17    & 50.31  & 65.80    & 66.13 & 63.49      & \textbf{71.32} \\
                    31    & 50.31  & 64.95    & 62.48 & 68.41      & \textbf{72.14} \\
                    50    & 50.31  & 63.32    & 56.97 & 69.38      & \textbf{71.91} \\ \bottomrule
                    \end{tabular}
            \end{adjustbox}
        \end{minipage}
    \end{minipage}}
    \caption{Performance versus $k_1$ on MSMT17.}
    \label{fig:K1_MSMT17}
\end{figure}

\begin{figure}[tb]
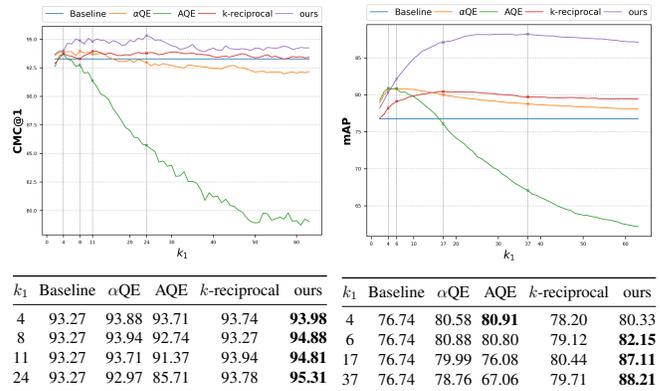

    \centering
    \subfloat[CMC@1 (\%) wrt. $k_1$]{
    \begin{minipage}[tb]{0.48\columnwidth}
        \begin{minipage}[tb]{\columnwidth}
            \centering
            \includegraphics[width=\columnwidth]{CMC1.png}
        \end{minipage}
        \vskip 0.15cm
        \begin{minipage}[tb]{\columnwidth}
            \centering
            \begin{adjustbox}{width=\columnwidth,center}
                \setlength\tabcolsep{3pt}
                \begin{tabular}{@{}cccccc@{}}
                    \toprule
                    $k_1$ & Baseline & $\alpha$QE & AQE     & $k$-reciprocal & ours             \\ \midrule
                    4     & 93.27  & 93.88    & 93.71 & 93.74      & \textbf{93.98} \\
                    8     & 93.27  & 93.94    & 92.74 & 93.27      & \textbf{94.88} \\
                    11    & 93.27  & 93.71    & 91.37 & 93.94      & \textbf{94.81} \\
                    24    & 93.27  & 92.97    & 85.71 & 93.78      & \textbf{95.31} \\ \bottomrule
                    \end{tabular}
            \end{adjustbox}
        \end{minipage}
    \end{minipage}}
    \subfloat[mAP (\%) wrt. $k_1$]{
    \begin{minipage}[tb]{0.48\columnwidth}
        \begin{minipage}[tb]{\columnwidth}
            \centering
            \includegraphics[width=\columnwidth]{mAP.png}
        \end{minipage}
        \vskip 0.15cm
        \begin{minipage}[tb]{\columnwidth}
            \centering
            \begin{adjustbox}{width=\columnwidth,center}
                \setlength\tabcolsep{3pt}
                \begin{tabular}{@{}cccccc@{}}
                    \toprule
                    $k_1$ & Baseline & $\alpha$QE & AQE              & $k$-reciprocal & ours             \\ \midrule
                    4     & 76.74  & 80.58    & \textbf{80.91} & 78.20      & 80.33          \\
                    6     & 76.74  & 80.88    & 80.80          & 79.12      & \textbf{82.15} \\
                    17    & 76.74  & 79.99    & 76.08          & 80.44      & \textbf{87.11} \\
                    37    & 76.74  & 78.76    & 67.06          & 79.71      & \textbf{88.21} \\ \bottomrule
                    \end{tabular}
            \end{adjustbox}
        \end{minipage}
    \end{minipage}}
    \caption{Performance versus $k_1$ on the small subset of VERI-Wild.}
    \label{fig:K1_SmallVeRiWild}
\end{figure}

\textbf{Neighborhood size $\mathbf{k_1}$.}
We use $k_1$ to control the number of nearest neighbors.
Generally, larger $k_1$ means richer contextual information but also involves more false matches.
The previous section tests the performance with empirically decided parameters, which may not fully reveal the maximum capacity for each method.
For example, $\alpha$QE can leverage contextual cues in broader neighborhoods than AQE because of the power-normalized combination weight. 
To better study the influence of neighborhood size, we evaluate re-ranking methods via gradually increasing $k_1$.
The performance variation is compared amongst AQE, $\alpha$QE, $k$-reciprocal re-ranking, and our method.
We fix the $\alpha$ in $\alpha$QE as 3 and $\lambda$ in $k$-reciprocal re-ranking as 0.3.
Because of the space limitation, we only show the figures for MSMT17 in \cref{fig:K1_MSMT17} and the small subset of VERI-Wild in \cref{fig:K1_SmallVeRiWild}.
The $k_2$ is fixed as 6 for both datasets.

\graphicspath{ {SSD/performance/MSMT17/comparison/} }

\graphicspath{ {SSD/performance/SmallVeRiWild/comparison/} }

We list the peak performance below in a table with the best result in each row marked with bold type.
From the curve plots, we have the following observations.
First, the performance of AQE drops quickly as $k_1$ becomes larger because AQE assigns uniform weight to each neighbor.
The aggregated feature embedding will be pulled towards the wrong directions if false matches dominate the neighborhood.
The $\alpha$QE uses power-normalized combination weights, which significantly relieves the false-match pollution issue. 
However, it is still not enough to entirely reject falsely retrieved samples resulting in the gradual performance degeneration. 
Second, our approach consistently outperforms $\alpha$QE for CMC@1 and mAP with a large margin.
Given a specific $k_1$, the performance gap proves that the correlation prediction provides a more accurate direction to shrink embeddings toward their identity centers.
Third, compared to $\alpha$QE, the $k$-reciprocal rule with backward verification is more robust to false matches.
The turning point of $k$-reciprocal re-ranking where the performance begins to drop comes later than $\alpha$QE.

\begin{figure}[tb]
    \centering
    \subfloat[mAP (\%) wrt. $k_2$]{
    \begin{minipage}[tb]{0.96\columnwidth}
        \begin{minipage}[tb]{\columnwidth}
            \centering
            \includegraphics[width=\columnwidth]{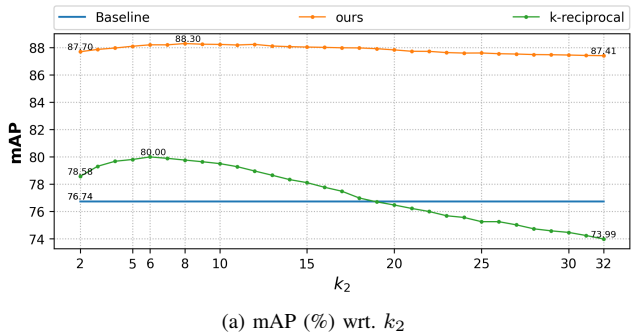}
        \end{minipage}
    \end{minipage}}
    \caption{Performance variation versus $k_2$ for k-reciprocal re-ranking and our method.}
    \label{fig:K2_SmallVeRiWild}
\end{figure}

\textbf{Refinement sequence $\mathbf{k_2}$.}
As mentioned above, memory refinement reduces feature pollution from falsely aggregated neighbors using a small group of high-confident samples. 
To measure the sensitivity, we test the performance of our method on the small subset of VERI-Wild with different $k_2$.
\cref{fig:K2_SmallVeRiWild} shows the variation with $k_1$ fixed as 32.

\graphicspath{ {SSD/performance/search_k2/SmallVeRiWild/} }

The result shows that mAP increases as the refinement sequence becomes longer, starting from a small value.
We reach the best mAP when refining on the 8-nearest neighbors.
After that, mAP gradually drops if we continue to include more samples.
Compared to the best reuslt 88.30\%, mAP drops to 87.41\% when $k_2=32$, which is even worse than $k_2=2$ whose mAP reaches 87.70\%.
This indicates that a small group of highly-confident neighboring embeddings are helpful in improving the discriminability of memory.
We also plot the performance of $k$-reciprocal re-ranking versus $k_2$ in \cref{fig:K2_SmallVeRiWild} which controls the local query expansion.
The decrease of mAP shows that it is more sensitive to $k_2$ because of its indiscriminate combination.

\subsection{Model Studies}
In this section, we study the influence of model architectures.
Specifically, we verify the effectiveness of the proposed modules in \cref{sec:Ablation} by conducting ablation experiments.
Besides, we test how the model architecture parameters, such as the number of heads in MHA, affect the final result in \cref{sec:Model Architecture Parameters}.

\subsubsection{Ablation} \label{sec:Ablation}
The proposed model comprises multiple modules, i.e., multi-block feature fusion, BaseEncoder, Contextual Memory, and memory refinement.
We train the model utile convergence with one or multiple modules removed. 
The re-ranking performance is tested on the small subset of VERI-Wild with $k_1=25, k_2=6$.
Results are shown in \cref{table:ablation study}.
Note that the output from BaseEncoder will be fed into the final binary classifier if the memory cell is disabled.

\begin{table}[]
    \caption{Ablation study. In the first row, experiments are tagged with Exp-$X$ where $X$ is a capital letter.
    The best and second best results are marked in red and blue respectively.}
    \begin{adjustbox}{width=\columnwidth,center}
        \setlength\tabcolsep{3pt}
        \begin{tabular}{@{}c|cccccccc@{}}
            \toprule
            Modules                    & Baseline & Exp-A   & Exp-B   & Exp-C   & Exp-D   & Exp-E                          & Exp-F   & Exp-G                          \\ \midrule
            Multi-Block Feature Fusion &          &         & \checkmark       & \checkmark       &         &                                & \checkmark       & \checkmark                              \\
            BaseEncoder                &          & \checkmark       & \checkmark       & \checkmark       &         & \checkmark                              &         & \checkmark                              \\
            Memory cell                &          &         &         & \checkmark       & \checkmark       & \checkmark                              & \checkmark       & \checkmark                              \\
            Memory refinement          &          &         &         &         & \checkmark       & \checkmark                              & \checkmark       & \checkmark                              \\ \midrule
            CMC@1                      & 93.27\%  & 86.18\% & 86.98\% & 92.87\% & 92.07\% & {\color[HTML]{3166FF} 94.38\%} & 93.37\% & {\color[HTML]{FE0000} 95.18\%} \\
            mAP                        & 76.74\%  & 77.93\% & 77.90\% & 86.31\% & 84.20\% & {\color[HTML]{3166FF} 87.09\%} & 86.56\% & {\color[HTML]{FE0000} 88.16\%} \\ \bottomrule
            \end{tabular}
    \end{adjustbox}
    \label{table:ablation study}
\end{table}

\textbf{Multi-Block Feature Fusion}.
We conducted two experiments (Exp-E and Exp-G) to study whether features from shallower layers contain discriminative information.
Results show the CMC@1 and mAP improve from 94.38\% to 95.18\% and 87.09\% to 88.16\%, respectively. 
The improvement verifies our assumption that different blocks provide information in various granularities.
Our re-ranking approach can make use of this information that is discarded in the baseline model.

\textbf{BaseEncoder}.
Before memory initialization, multi-block features are preprocessed by the BaseEncoder, which aggregates the contextual information for each embedding.
Comparing Exp-F and Exp-G, we can observe that CMC@1 and mAP drop 1.81\% and 1.60\%, respectively, with BaseEncoder disabled.
However, the decline is more severe if multi-block feature fusion is removed at the same time.
From Exp-E to Exp-D, the CMC@1 and mAP decrease 2.31\% and 2.89\%, respectively. 
A possible explanation is that the multi-block feature vector provides richer information than the baseline.
Once discriminative features from shallower blocks are cut off, the ability of BaseEncoder to aggregate homogeneous sub-features plays an important role.

\textbf{Contextual Memroy} is the most critical module in our architecture.
As shown in Exp-B and Exp-C, directly predicting correlations for embeddings generated by BaseEncoder significantly harms the performance.
The CMC@1 decreases from 92.87\% to 86.98\% with 5.89\% decline.
Similarly, mAP also drops 8.41\%, reaching 77.90\%.
The performance drop suggests that a fixed hyperplane is not enough to separate embeddings from Transformer encoder.
Instead, the similarity between each embedding counts.

\begin{figure}[tb]
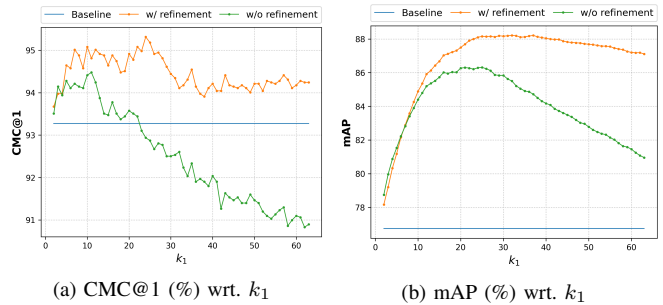

    \centering
    \subfloat[CMC@1 (\%) wrt. $k_1$]{
    \begin{minipage}[tb]{0.48\columnwidth}
        \begin{minipage}[tb]{\columnwidth}
            \centering
            \includegraphics[width=\columnwidth]{CMC1.png}
        \end{minipage}
    \end{minipage}}
    \centering
    \subfloat[mAP (\%) wrt. $k_1$]{
    \begin{minipage}[tb]{0.49\columnwidth}
        \begin{minipage}[tb]{\columnwidth}
            \centering
            \includegraphics[width=\columnwidth]{mAP.png}
        \end{minipage}
    \end{minipage}}
    \caption{Ablation study with memory refinement module removed.}
    \label{fig:ablation refinement}

\end{figure}

\textbf{Memory refinement} is a sub-module of the Contextual Memory, which aims at preventing feature pollutions.
We conduct Exp-C and Exp-G to study the refinement process with $k_1$ fixed as $25$.
The CMC@1 increases from 92.87\% to 95.18\% with 2.31\% improvement, and mAP also improved 1.85\%.
To better reveal the power of memory refinement, we evaluate the re-ranking performance under different $k_1$ for the consideration that refinement is more crucial if more false matches are included in the neighborhood.
The metric curve versus $k_1$ is shown in \cref{fig:ablation refinement}.
Comparing the results, we can observe that the model without refinement deteriorates rapidly when $k_1$ becomes larger.
Memory refinement enables our model to leverage contextual information in larger neighborhoods without being affected by false matches.
\graphicspath{ {SSD/performance/refinement/SmallVeRiWild/} }

\subsubsection{Model Architecture Parameters} \label{sec:Model Architecture Parameters}
The ablation studies verify the effectiveness of each proposed module. 
Here, we perform experiments to study the relationships between architecture parameters and the re-ranking performance.
The basic model follows Exp-G in \cref{table:ablation study} with all the modules enabled.
Each time, we set different values for an architecture parameter with others fixed.
After the model converges, CMC@1 and mAP are tested on the small subset of VERI-Wild with $k_1=25, k_2=6$.

\textbf{BaseEncoder Layers.}
We control the model capacity by adjusting the number of layers in the BaseEncoder.
Results are shown in \cref{table:BaseEncoder_Layers}.
We can observe that the two-layer BaseEncoder obtains the best results.
If we stack more layers, the performance gradually decreases but is relatively stable.
At the same time, we noticed that the model becomes harder to converge.
The warm-up period is doubled to 20 epochs to stabilize the model at the initial training stages for 4-layer, 5-layer, and 6-layer BaseEncoder.
The model degeneration is likely to be caused by that deeper BaseEncoder affects the backflow of gradient.
\begin{table}[]
    \caption{Performance variation wrt. number of encoder layers in the BaseEncoder.}
        \setlength\tabcolsep{3pt}
        \centering
        \begin{tabular}{@{}c|cccccc@{}}
            \toprule
            Layers & 1 & 2 & 3 & 4 & 5 & 6 \\ \midrule
            CMC@1(\%) & 94.14 & 95.18 & 95.18 & 94.78 & 94.65 & 94.31 \\
            mAP(\%) & 87.22 & 88.21 & 88.16 & 88.14 & 88.10 & 87.75 \\ \bottomrule
        \end{tabular}
    \label{table:BaseEncoder_Layers}
\end{table}

\textbf{Heads in Multi-Head Attention.} MHA is one of the important building blocks in our model.
The head refers to splitting original feature embeddings into multiple smaller ones for attention calculation independently.
As shown in \cref{table:heads}, results are unsatisfying for models with limited number of heads.
For example, the CMC@1 drops from 95.18\% to 92.03\% when heads reduce from 16 to 1.
The mAP also drops 3.88\%.
Intuitively, the multi-head structure captures the similarity between different sub-features.
It can be considered an improved version of \cite{yu2017divide} where sub-features are selected either manually or randomly.
The projection matrices in Multi-Head Attention are learned from data, which enable the transformed sub-features to focus on the most discriminative regions.
\begin{table}[]
    \caption{Performance variation wrt. number of heads in MHA.}
    \begin{adjustbox}{width=\columnwidth,center}
        \setlength\tabcolsep{3pt}
        \begin{tabular}{@{}c|cccccccc@{}}
            \toprule
            Heads & 1 & 2 & 4 & 8 & 16 & 32 & 64 & 128 \\ \midrule
            CMC@1(\%) & 92.03 & 92.00 & 93.47 & 94.24 & 95.18 & 94.61 & 94.61 & 94.54 \\
            mAP(\%) & 84.28 & 83.77 & 85.98 & 87.01 & 88.16 & 87.97 & 88.02 & 87.94 \\ \bottomrule
        \end{tabular}
    \end{adjustbox}
    \label{table:heads}
\end{table}

\textbf{Memory Size} refers to the number of slots in the Contextual Memory.
Similar to the Multi-Head Attention, each memory slot focuses on some specific aspects.
Therefore, the memory size relates to the distinctive aspects of a person or vehicle.
We study the memory size in \cref{table:memory_size}.
From the table, the eight-slot memory outperforms others.
An interesting finding is that the number 8 happens to be the same as the views in multi-view feature inference \cite{zhou2018vehicle}, which suggests the memory cell stores view-related information inside.
Further increasing the memory size slightly harms the performance.
However, the affect is more significant if we adpot a smaller memory.
The CMC@1 decreases 1.47\% and mAP drops 1.73\% from eight-slot memory to tow-slot memory because small Contextual Memory can not provide enough discriminative features for neighborhood reconstruction.
\begin{table}[]
    \caption{Performance variation wrt. the memory size.}
    \begin{adjustbox}{width=\columnwidth,center}
        \setlength\tabcolsep{3pt}
        \begin{tabular}{@{}c|ccccccccc@{}}
        \toprule
            Memory Size & 2 & 4 & 8 & 12 & 16 & 20 & 24 & 28 & 32 \\ \midrule
            CMC@1(\%) & 93.71 & 94.51 & 95.18 & 94.98 & 94.95 & 94.88 & 94.48 & 94.28 & 94.68 \\
            mAP(\%) & 86.43 & 87.51 & 88.16 & 87.98 & 87.87 & 87.88 & 87.69 & 87.49 & 87.52 \\ \bottomrule
        \end{tabular}
    \end{adjustbox}
    \label{table:memory_size}
\end{table}

\section{Conclusion}
In this paper, we have proposed a novel Contextual Memory cell to mimic the remembering process that humans adopt for re-ID and re-ranking.
By comparing neighbors' features with the multi-view appearance information in the memory, we predict the correlations between each image and its $k$-nearest neighbors.
The re-ranking is achieved by shrinking each embedding towards the identity centers with correlation prediction as discriminative combination weights.
Experiments on six widely-used re-ID datasets validate the effectiveness of the proposed method.
Especially, the performance boost on large-scale datasets VERI-Wild, MSMT17, and VehicleID exhibits the ability of ACP in handling complex neighborhood relationships.


%


\ifCLASSOPTIONcaptionsoff
  \newpage
\fi

\bibliography{egbib.bib}

%






\end{document}